\documentclass[twoside,11pt]{article}

\usepackage{blindtext}

\usepackage{jmlr2e}

\usepackage{amsmath,amssymb,amsfonts}
\usepackage{algorithmic}
\allowdisplaybreaks[4]
\usepackage{svg}
\usepackage{booktabs}
\usepackage{threeparttable}
\usepackage{tablefootnote}
\usepackage{url}
\usepackage{multirow}
\usepackage[algoruled,linesnumbered]{algorithm2e}
\usepackage{cite}
\usepackage{tikz}
\usepackage{graphicx}
\usepackage{hyperref}
\usepackage{tabularx}
\usepackage{array}
\usepackage{booktabs}
\usepackage{verbatim}
\usepackage{subfigure}
\usepackage[labelfont=bf,textfont={bf}]{caption}
\usepackage{xcolor}
\usepackage{appendix}
\usepackage{ragged2e}
\usepackage{bm}
\usepackage{float}

\usepackage{lastpage}
\jmlrheading{xx}{xxxx}{1-\pageref{LastPage}}{7/24; Revised 5/25}{x/xx}{xx-xxxx}{Xianghua Zeng, Hao Peng, Dingli Su, and Angsheng Li}

\ShortHeadings{Hierarchical Decision Making Based on Structural Information Principles}{Zeng, Peng, Su, AND Li}
\firstpageno{1}

\begin{document}

\title{Hierarchical Decision Making Based on Structural Information Principles}

\author{\name Xianghua Zeng \email zengxianghua@buaa.edu.cn \\
       \addr School of Computer Science and Engineering\\
       Beihang University\\
       Beijing, 100191, China
       \AND
       \name Hao Peng \email penghao@buaa.edu.cn \\
       \addr School of Cyber Science and Technology\\
       Beihang University\\
       Beijing, 100191, China
       \AND
       \name Dingli Su \email sudingli@buaa.edu.cn \\
       \addr School of Computer Science and Engineering\\
       Beihang University\\
       Beijing, 100191, China
       \AND
       \name Angsheng Li \email angsheng@buaa.edu.cn \\
       \addr School of Computer Science and Engineering\\
       Beihang University\\
       Beijing, 100191, China
       }

\editor{My editor}

\maketitle

\begin{abstract}
Hierarchical Reinforcement Learning (HRL) is a promising approach for managing task complexity across multiple levels of abstraction and accelerating long-horizon agent exploration.
However, the effectiveness of hierarchical policies heavily depends on prior knowledge and manual assumptions about skill definitions and task decomposition.
In this paper, we propose a novel \textbf{S}tructural \textbf{I}nformation principles-based framework, namely \textbf{SIDM}, for hierarchical \textbf{D}ecision \textbf{M}aking in both single-agent and multi-agent scenarios.
Central to our work is the utilization of structural information embedded in the decision-making process to adaptively and dynamically discover and learn hierarchical policies through environmental abstractions.
Specifically, we present an abstraction mechanism that processes historical state-action trajectories to construct abstract representations of states and actions.
We define and optimize directed structural entropy—a metric quantifying the uncertainty in transition dynamics between abstract states—to discover skills that capture key transition patterns in RL environments.
Building on these findings, we develop a skill-based learning method for single-agent scenarios and a role-based collaboration method for multi-agent scenarios, both of which can flexibly integrate various underlying algorithms for enhanced performance.
Extensive evaluations on challenging benchmarks demonstrate that our framework significantly and consistently outperforms state-of-the-art baselines, improving the effectiveness, efficiency, and stability of policy learning by up to $32.70\%$, $64.86\%$, and $88.26\%$, respectively, as measured by average rewards, convergence timesteps, and standard deviations.
\end{abstract}

\begin{keywords}
  Hierarchical reinforcement learning, Structural information principles, State and action abstractions, Skill-based learning, Role-based learning
\end{keywords}

\section{Introduction} \label{sec: introduction}
Reinforcement learning (RL) \citep{sutton1998introduction} enables agents to develop optimal decision-making strategies by interacting with their environment to solve sequential tasks with goal-oriented objectives.
The integration of deep neural networks \citep{lecun2015deep, schmidhuber2015deep} with RL has demonstrated remarkable success in diverse applications, including game intelligence \citep{AlphaStar2019, zhou2023malib, zhong2024heterogeneous}, video acceleration \citep{ramos2022text}, and model fitting \citep{truong2022unsupervised}.
However, RL algorithms face the critical challenge of requiring extensive interactions with complex environments to learn effective policies \citep{matteshieros}.

Hierarchical reinforcement learning (HRL) offers a promising approach to improving sample efficiency by structuring agent exploration and decision-making across multiple levels of abstraction \citep{merel2018lhierarchical, marino2018hierarchical}. 
HRL decomposes long-horizon tasks into subtasks, enabling hierarchical policies to operate at different temporal and abstraction levels, thereby accelerating policy learning \citep{hafner2022deep}. 
However, many existing HRL approaches depend on prior knowledge, incorporating handcrafted assumptions about skill definitions \citep{lee2019composing, lee2020learning} or manually designed task decomposition heuristics \citep{tessler2017deep}.
The HSD-3 framework \citep{gehring2021hierarchical} autonomously learns a skill hierarchy during a pre-training stage without reward supervision, but it still relies on manually selected features and predefined target objectives.
Reskill \citep{rana2023residual} leverages state-conditioned generative models to construct a skill space but depends on expert-defined manipulation tasks for collecting demonstration data.
The CEO framework \citep{machado2023temporal}, built on successor representation \citep{dayan1993improving}, effectively discovers meaningful skills but introduces computational complexity and is sensitive to the skill scale.
In multi-agent reinforcement learning (MARL), role-based task decomposition \citep{wang2020roma, wang2020rode} has proven effective in facilitating hierarchical collaborative strategies. 
However, its success heavily depends on domain-specific task knowledge and is highly sensitive to role discovery parameters.
Therefore, developing an effective and stable hierarchical decision-making framework that operates without prior knowledge remains a critical challenge in advancing scalable and generalizable reinforcement learning.

To address this, we draw inspiration from structural information principles \citep{li2016structural}, which provide a foundation for adaptive hierarchy discovery.
Structural entropy measures the uncertainty in an undirected graph's dynamics by quantifying the number of bits required to encode a vertex transition during a single-step random walk.
Minimizing this structural uncertainty yields a hierarchical partitioning of graph vertices\footnote{A vertex is defined in the graph and a node in the tree.}, referred to as the encoding tree.
Within this tree, each node corresponds to a subset of vertices, termed a 'community', where vertices exhibit stronger intraconnections than interconnections.
In this work, we leverage the structural information embedded in the decision-making process to discover and learn hierarchical policies dynamically through environmental abstractions.

To this end, we propose a novel \textbf{S}tructural \textbf{I}nformation principles-based hierarchical \textbf{D}ecision \textbf{M}aking framework, called \textbf{SIDM}, to address the reliance on prior knowledge and manual design in HRL.
First, we introduce an adaptive abstraction mechanism that extracts meaningful structure from high-dimensional, noisy state-action information, producing compact abstract representations of states and actions. 
This mechanism leverages encoding trees to cluster states or actions with similar features into communities dynamically and applies an aggregation function to compute their abstract representations.
Second, we formally define and optimize directed structural entropy to extend structural information principles beyond undirected graphs, enabling the modeling of asymmetric abstract state transitions. 
Using directed entropy, we quantify transition probabilities between abstract states and identify high-frequency transitions between abstract communities as skills, capturing key transition patterns in RL environments.
Third, we build on the discovered skills and abstract actions to develop a skill-based method for single-agent learning and a role-based strategy for multi-agent collaboration.
These methods operate independently of manual assistance and can flexibly integrate various underlying algorithms to enhance their performance.
Finally, we conduct extensive experiments and comprehensive analysis on well-established benchmarks, including visual gridworld navigation, continuous robotic control, and StarCraft II micro-management. 
Comparative results demonstrate that, compared to state-of-the-art baselines, our framework improves average reward (effectiveness) and sampling efficiency (efficiency) by up to $32.70\%$ and $64.86\%$, respectively, while reducing standard deviation (stability) by up to $88.26\%$.
% To facilitate reproducibility and further research, we provide the source code and demonstration videos for SIDM on GitHub\footnote{\url{https://github.com/SELGroup/SIDM}}.

\begin{figure}[t]
    \centering
    \includegraphics[width=\textwidth]{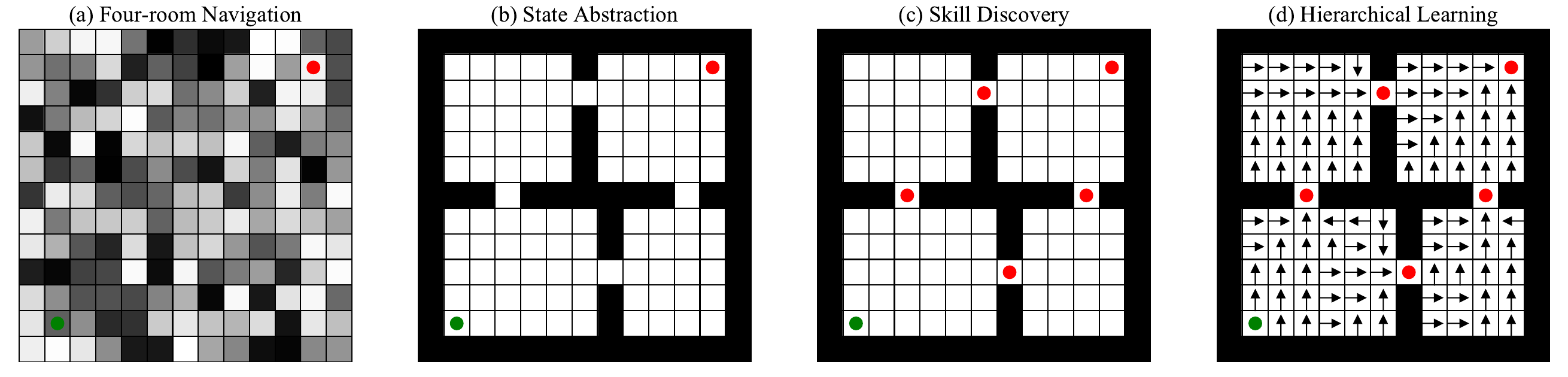}
    \vspace{-0.7cm}
    \caption{Illustrative single-agent navigation task within the gridworld benchmark, where the agent navigates from the start location (green) to the goal (red) while interacting with its environment.} \label{fig: example}
    \vspace{-0.7cm}
\end{figure}

Figure \ref{fig: example} shows a single-agent navigation task in the four-room domain, where the agent moves between rooms to reach a designated target.
As shown in Figure \ref{fig: example}(a), the agent receives high-dimensional, noisy visual inputs (e.g., from a camera or sensor) and navigates from the green starting position to the red target location.
As illustrated in Figure \ref{fig: example}(b), the SIDM framework extracts structural relationships from raw observations to generate abstract state representations in a lower-dimensional space, approximating the original 2D coordinates and thereby simplifying decision-making.
By computing and optimizing structural entropy in directed abstract transitions, the SIDM framework adaptively identifies key abstract states—such as turning points between rooms in the gridworld (Figure \ref{fig: example}(c)).
By modeling navigation behaviors between these key states as skills, SIDM enables more efficient decision-making within each corresponding subspace, as shown in Figure \ref{fig: example}(d).

This paper is organized as follows: Section \ref{section: preliminaries} outlines the preliminaries and notations, Section \ref{section: related work} discusses the related work, Section \ref{section: framework} presents the detailed designs of SIDM framework, Sections \ref{section: setup} and \ref{section: evaluation} describe the experimental setups and evaluations, followed by the conclusion in Section \ref{section: conclusion}.

% Compared to our preliminary works presented in the proceedings of AAAI \citep{zeng2023effective} and IJCAI \citep{zeng2023hierarchical}, this journal version includes significant enhancements in both methodology and structure.
% Specifically, we combine the abstractions for single-agent states and multi-agent actions into a unified mechanism that employs the new stretch and compress operators \citep{pan2021information}, replacing the original optimization operators \citep{li2018decoding} to enhance performance.
% To measure the dynamic uncertainty in directed transitions between abstract states, we extend the current undirected structural information \citep{li2016structural} to formally define and optimize directed entropy for unsupervised skill discovery.
% Building on the adaptive abstraction mechanism and unsupervised skill discovery, we develop a skill-based single-agent learning method and a role-based multi-agent collaboration method, enriching the framework's structure.
% In the experimental section, we consider advanced baselines and benchmarks to highlight the performance advantages of our framework.

\section{Preliminaries and Notations} \label{section: preliminaries}
This section establishes the foundational concepts of reinforcement learning and structural information principles.
We distinguish between ``primitive states and actions,'' which represent the original variables provided by the environment, and ``abstract states and actions,'' which represent higher-level abstractions derived from them.
Spaces (denoted by calligraphic fonts, e.g., $\mathcal{S}, \mathcal{A}$) represent the complete sets of possible states or actions.
Variables (denoted by capital letters, e.g., $S, A$) represent random variables over states or actions, often sampled from the replay buffer in practical implementations.
Values (denoted by lowercase letters, e.g., $s, a$) represent specific instances of states or actions within their respective spaces.
A summary of the primary notations and their detailed descriptions is provided in Appendix \ref{app: notations}.

\subsection{Reinforcement Learning}
Reinforcement learning (RL) is a learning paradigm in which one or more agents learn to make sequential decisions by interacting with an environment whose dynamics may be partially or fully unknown, with the objective of maximizing expected cumulative rewards.

\subsubsection{Markov Decision Process}
In RL, the single-agent decision-making problem is formulated as a \textbf{Markov decision process} (MDP) \citep{bellman1957markovian}, which is formally defined as a tuple:
\begin{equation}
    \mathcal{M}_s = \langle \mathcal{S}, \mathcal{A}, \mathcal{R}, \mathcal{P}, \gamma \rangle\text{,}
\end{equation}
where $\mathcal{S}$ is the state space, $\mathcal{A}$ is the action space, $\mathcal{R}: \mathcal{S} \times \mathcal{A} \rightarrow \mathbb{R}$ is the reward function mapping state-action pairs to expected rewards, $\mathcal{P}: \mathcal{S} \times \mathcal{A} \rightarrow \Delta(\mathcal{S})$ is the transition function defining the probability distribution over next states, and $\gamma \in [0,1)$ is the discount factor that determines the weighting of future rewards.

The notation $\Delta(\mathcal{S})$ refers to the space of probability distributions over the state space $\mathcal{S}$. Here, the subscript $s$ in $\mathcal{M}_s$ explicitly denotes a single-agent MDP, distinguishing it from the multi-agent Markov game $\mathcal{M}_m$ introduced later.

At each timestep $t$, the agent observes the current state $s_t \in \mathcal{S}$ and selects an action $a_t \in \mathcal{A}$ according to its policy, i.e., $a_t \sim \pi(s_t)$, where $\pi: \mathcal{S} \rightarrow \Delta(\mathcal{A})$ defines a probability distribution over actions conditioned on the current state. 
The action $a_t$ leads to a new state $s_{t+1}$, sampled from the transition distribution, i.e., $s_{t+1} \sim \mathcal{P}(s_t, a_t)$, and the agent receives a reward $r_t \sim \mathcal{R}(s_t, a_t) \in \mathbb{R}$.

The agent aims to learn an optimal policy $\pi^*: \mathcal{S} \rightarrow \Delta(\mathcal{A})$ that maximizes the expected cumulative discounted reward:
\begin{equation}
    \pi^* = \arg\max _\pi \mathbb{E}_{\mathcal{P}, \pi}\left[\sum_{t=0}^{\infty} \gamma^t \mathcal{R}(s_t, a_t)\right]\text{.}
\end{equation}

\subsubsection{Markov Game}
A fully cooperative multi-agent task, where all agents aim to maximize a shared reward, is modeled as a \textbf{Markov game} \citep{littman1994markov}. 
It is formally defined as a tuple:
\begin{equation}
    \mathcal{M}_m = \langle \mathcal{N}, \mathcal{S}, \mathcal{A}, \mathcal{R}, \mathcal{P}, \gamma \rangle\text{,}
\end{equation}
where $\mathcal{N} \equiv \{n_1, n_2, \ldots, n_{|\mathcal{N}|}\}$ denotes the finite set of cooperative agents, with $|\mathcal{N}|$ representing the total number of agents. 
$\mathcal{S}$ is the global state space, and $\mathcal{A} \equiv \mathcal{A}_1 \times \cdots \times \mathcal{A}_{|\mathcal{N}|}$ is the joint action space composed of individual agent action spaces. 
$\mathcal{R}: \mathcal{S} \times \mathcal{A} \rightarrow \mathbb{R}$ is the shared reward function, $\mathcal{P}: \mathcal{S} \times \mathcal{A} \rightarrow \Delta(\mathcal{S})$ is the global transition function, and $\gamma \in [0,1)$ is the discount factor.

At each timestep $t$, each agent $n_i \in \mathcal{N}$ selects an action $a_t^i \in \mathcal{A}_i$ based on the global state $s_t \in \mathcal{S}$. 
The set of individual actions forms a joint action $\bm{a}_t \equiv [a_t^i]_{i=1}^{|\mathcal{N}|}$. 
This joint action induces a transition to the next global state $s_{t+1} \sim \mathcal{P}(\cdot \mid s_t, \bm{a}_t)$ and yields a shared reward $r_t = \mathcal{R}(s_t, \bm{a}_t)$.

Each agent $n_i$ maintains a local trajectory $\tau_i$, which consists of its sequence of observations, actions, and rewards over time. 
It optimizes a local policy $\pi_i(a_t^i \mid \tau_i)$ to maximize the overall team performance, defined by the expected cumulative discounted reward under the joint policy $\bm{\pi} \equiv \{\pi_1, \pi_2, \ldots, \pi_{|\mathcal{N}|}\}$:
\begin{equation}
    \bm{\pi}^* = \arg\max _{\bm{\pi}} \mathbb{E}_{\mathcal{P}, \bm{\pi}}\left[\sum_{t=0}^{\infty} \gamma^t \mathcal{R}(s_t, \bm{a}_t)\right]\text{.}
\end{equation}

\subsubsection{State or Action Abstraction}
\textbf{State or action abstraction} \citep{abel2022theory} aims to simplify decision-making by designing a parameterized abstraction function $f_{\phi}$ with the trainable parameter $\phi$, which maps primitive states and actions to their abstract counterparts.

Specifically, $f_{\phi}$ maps each primitive state $s \in \mathcal{S}$ to an abstract state $z_s \in \mathcal{Z}_s$, i.e., $f_\phi: \mathcal{S} \rightarrow \mathcal{Z}_s$, or each primitive action $a \in \mathcal{A}$ to an abstract action $z_a \in \mathcal{Z}_a$, i.e., $f_\phi: \mathcal{A} \rightarrow \mathcal{Z}_a$.
This abstraction reduces the complexity of the original decision process by compressing state and action representations, enabling more efficient policy learning through reduced dimensionality and improved generalization.
The resulting abstract MDP is formally defined as a tuple:
\begin{equation}
    \mathcal{M}_\phi = \langle \mathcal{Z}_s, \mathcal{Z}_a, \mathcal{R}_\phi, \mathcal{P}_\phi, \gamma \rangle\text{,}
\end{equation}
where $\mathcal{Z}_s$ and $\mathcal{Z}_a$ are the abstract state and action spaces, $\mathcal{P}_\phi: \mathcal{Z}_s \times \mathcal{Z}_a \rightarrow \Delta(\mathcal{Z}_s)$ is the abstract transition function, $\mathcal{R}_\phi: \mathcal{Z}_s \times \mathcal{Z}_a \rightarrow \mathbb{R}$ is the abstract reward function, and $\gamma \in [0,1)$ is the discount factor.

The abstract transition and reward functions, $\mathcal{P}_\phi$ and $\mathcal{R}_\phi$, are constructed by aggregating the transition dynamics and reward distributions of the underlying MDP over the pre-image sets induced by $f_\phi$.
Specifically, they are defined as follows:
\begin{equation}
    \mathcal{P}_\phi(z^s_k|z^s_i,z^a_j) = \sum_{s_t \in z^s_i}\sum_{a_t \in z^a_j} \sum_{s_{t+1} \in z^s_k}\mathcal{P}(s_{t+1}|s_t,a_t)\text{,}\quad  \mathcal{R}_\phi(z^s_i,z^a_j) = \sum_{s_t \in z^s_i}\sum_{a_t \in z^a_j} \mathcal{R}(s_t,a_t)\text{,}
\end{equation}
where $z^s_i \in \mathcal{Z}_s$ and $z^a_j \in \mathcal{Z}_a$ denote an abstract state and abstract action, respectively.

\subsubsection{Skill-based Learning}
In \textbf{skill-based learning}, we use the term `skill' to refer broadly to the general concept of reusable behaviors or abilities, while an `option' refers to a specific, parameterized instance of a skill, defined by an initiation set, an option policy, and a termination condition, as established in prior work \citep{sutton1999between, machado2023temporal}.

Following the option framework \citep{precup2000temporal}, an option $\kappa \in \mathcal{K}$, representing a learned skill, is formally defined as a tuple:
\begin{equation}
    \kappa = \langle \mathcal{I}_\kappa, \pi_\kappa, \mathcal{T}_\kappa \rangle\text{,}
\end{equation}
where $\mathcal{I}_\kappa \subseteq \mathcal{S}$ is the initiation set in which the option $\kappa$ can be executed, $\pi_\kappa: \mathcal{S} \rightarrow \Delta(\mathcal{A})$ is is the option policy mapping states to a probability distribution over primitive actions, and $\mathcal{T}_\kappa: \mathcal{S} \rightarrow [0,1]$ is the termination function specifying the probability of terminating the option at a given state.

Incorporating the skill space $\mathcal{K}$ within an MDP gives rise to a hierarchical two-level policy structure.
The high-level policy $\pi_k^h$ selects an option $\kappa \in \mathcal{K}$, while the low-level policy $\pi_k^l$ governs primitive action execution under the chosen option until the termination function $\mathcal{T}_\kappa$ signals the end of the option.

\subsubsection{Role-based Learning}
In \textbf{role-based learning} \citep{wilson2010bayesian, wang2020rode}, the goal is to improve agent coordination and enhance scalability in complex multi-agent cooperative tasks by decomposing $\mathcal{M}_m$ into subtasks through the assignment of specialized roles.
This decomposition is guided by a predefined role space $\Psi$, where each role imposes structural constraints on agent behavior by limiting its available actions.
By reducing ambiguity in role assignments, this approach facilitates more efficient learning and execution of cooperative behaviors.

Each role $\rho_{j} \in \Psi$ encapsulates a subtask and an associated policy, formally defined as:
\begin{equation}
    \rho_j = \left\langle t_{j}, \pi_{\rho_{j}}\right\rangle\text{,}
\end{equation}
where the subtask $t_{j} = \langle \mathcal{N}_{j}, \mathcal{S}, \mathcal{A}_{j}, \mathcal{R}, \mathcal{P}, \gamma\rangle$  is derived from the global task but restricted to a subset of agents $\mathcal{N}_i \subseteq \mathcal{N}$.
The role policy $\pi_{\rho_{j}}: \mathcal{S} \rightarrow \Delta(\mathcal{A}_{j})$ governs action selection by assigning a probability distribution over the actions available within subtask $t_j$.

Within each subtask $t_j$, agents in $\mathcal{N}_j$ operate in a constrained action subspace $\mathcal{A}_j \subseteq \mathcal{A}$, which minimizes action overlap across roles and promotes more structured and efficient role-based coordination.

\subsection{Structural Information Principles}
Efficient and robust decision-making requires abstracting raw state and action information by eliminating irrelevant details while preserving essential features.
To model structural relationships among states or actions, we construct a weighted, undirected graph $G=(V, E, W)$, built separately for each entity type.
In this graph, all vertices $V$ represent either states $S$ or actions $A$, and edges $E$ connect vertices that exhibit functional similarity, typically computed via feature-based similarity metrics.
The edge weights $W: E \rightarrow \mathbb{R}_{\geq 0}$ quantify the degree of similarity between connected vertices, and the degree of a vertex $v \in V$, denoted $d_v$, is defined as the sum of the weights of all its incident edges.

Building on this graphical representation, we introduce encoding trees \citep{li2016structural} to represent a hierarchical partitioning of the vertices based on their feature similarities.
Each level of the tree reflects a progressive grouping of elements, where lower levels correspond to fine-grained partitions that capture detailed distinctions, and higher levels represent broader aggregations that highlight overarching structural patterns. 
This hierarchical structure helps capture both local and global structural relationships among states or actions, enabling the model to recognize patterns of similarity and connectivity that influence decision-making processes.

We then leverage structural entropy—a measure of uncertainty associated with hierarchical partitioning—to quantitatively assess the quality of the induced hierarchy. 
Minimizing structural entropy guides partitioning to capture the intrinsic organization of the state or action space more faithfully.

\subsubsection{Encoding Tree}
Given the weighted, undirected graph $G$, we define an \textbf{encoding tree} $T$ that hierarchically partitions the set of states or actions according to their feature-driven similarities. 
This tree is rooted and satisfies the following properties:

$\bullet$ The root node $\lambda \in T$ corresponds to the entire vertex set $V$, meaning that $V_{\lambda} = V$, and it serves as the starting point for the hierarchical partitioning.

$\bullet$ Each leaf node $\nu \in T$ corresponds to a singleton set containing a single vertex $v \in V$, i.e., $V_{\nu} = {v}$, representing the finest level of partitioning in the tree.

$\bullet$ Each internal node $\alpha \in T$ (i.e., neither a root nor a leaf) corresponds to a vertex subset $V_{\alpha} \subseteq V$, grouping states or actions that are further partitioned in the tree.

$\bullet$ Each non-root node $\alpha \in T$ has a unique parent node, denoted as $\alpha^-$, from which $\alpha$ is directly descended in the tree structure.

$\bullet$ Each non-leaf node $\alpha \in T$ has exactly $L_{\alpha} \geq 2$ children, indexed from left to right as $\alpha_1, \alpha_2, \dots, \alpha_{L_\alpha}$, where the index $i$ increases sequentially.

$\bullet$ For each non-leaf node $\alpha \in T$, the vertex subsets of its children are pairwise disjoint, satisfying $V_{\alpha} = \bigcup_{i=0}^{L_\alpha-1} V_{\alpha_i}$ and $V_{\alpha_i} \cap V_{\alpha_j} = \emptyset$ for all $i \neq j$.

\subsubsection{Structural Entropy}
The \textbf{one-dimensional structural entropy} measures the dynamical uncertainty in single-step random walks within the state or action graph $G$, without applying any partitioning strategy, where all states or actions belong to a single community.
Mathematically, it is equivalent to the Shannon entropy of the stationary distribution induced by vertex degrees, reflecting the local similarity structure of each state or action with its neighbors.
It is defined as follows:
\begin{equation}\label{equ: 1d_se}
    H^{1}(G)=-\sum_{v \in V} \frac{d_{v}}{\operatorname{vol}(G)} \cdot \log _{2} \frac{d_{v}}{\operatorname{vol}(G)}\text{,}
\end{equation}
where $\operatorname{vol}(G)=\sum_{v \in V} d_{v}$ is the total volume of the graph, representing the sum of all vertex degrees.
An encoding tree $T$ introduces a hierarchical partitioning strategy that refines this entropy by clustering strongly connected vertices, thereby revealing the inherent hierarchical community structure of the graph.

The \textbf{\bm{$K$}-dimensional structural entropy} quantifies the residual structural entropy in the graph $G$ after applying a hierarchical partitioning strategy represented by an encoding tree $T$ with height at most $K$. 
For each non-root node $\alpha$ in $T$, the assigned structural entropy is defined as follows:
\begin{equation}\label{equ: kd_node_se}
    H^{T}(G;\alpha)=-\frac{g_{\alpha}}{\operatorname{vol}(G)} \log _{2} \frac{\mathcal{V}_{\alpha}}{\mathcal{V}_{\alpha^{-}}}\text{,}
\end{equation}
where $g_{\alpha}$ denotes the total edge weight of all edges crossing the boundary of $V_{\alpha}$, and $\mathcal{V}_{\alpha}$ denotes the volume of the subgraph induced by $V_{\alpha}$, i.e., the sum of degrees of its constituent vertices.
A higher value of $H^{T}(G;\alpha)$ indicates greater uncertainty in single-step transitions from the parent community $V_{\alpha^-}$ to the sub-community $V_\alpha$.

The overall $K$-dimensional structural entropy is defined as the minimum total entropy over all encoding trees of height at most $K$, formally expressed as follows:
\begin{equation} \label{equ: kd_se}
    H^{K}(G)=\min_{T}\left\{H^T(G)\right\}\text{,}\quad H^T(G)=\sum_{\alpha \in T, \alpha \neq \lambda}H^{T}(G;\alpha)\text{.}
\end{equation}
A structural entropy optimization algorithm guides the construction of such an encoding tree, and the tree that minimizes this entropy is referred to as the optimal encoding tree. 
This tree effectively captures the hierarchical community structure among states or actions based on feature similarity. 
In each sampled batch, states or actions with similar features are adaptively grouped into communities, forming the foundation for subsequent state and action abstraction mechanisms.

\section{Related Work} \label{section: related work}
This section reviews related work on structural information principles, state abstraction, and hierarchical decision-making, highlighting the motivation behind our proposed SIDM framework and its primary advantages over existing approaches.

\subsection{Structural Information Principle}
Structural information \citep{li2016structural} was introduced in 2016 to quantify the dynamic uncertainty in complex graph structures.
Specifically, structural entropy measures the minimum number of bits required to identify an accessible vertex in a single-step random walk.
The principles of structural entropy minimization \citep{li2016three, li2018decoding} were later introduced to automatically identify the optimal partitioning strategy, known as the encoding tree, which reveals the hierarchical community structure of vertices.

Since then, structural information principles have been applied across various fields, including graph learning \citep{liu2019rem, wu2022structural}, network analysis \citep{zeng2024adversarial, cao2024hierarchical}, and reinforcement learning \citep{zeng2023effective, zeng2023hierarchical}.
However, existing research on structural information has primarily focused on undirected graphs, which are unable to capture irreversible directional relationships between graph vertices, leading to inevitable information loss.

In this work, we extend structural information principles by proposing a formal definition and an optimization algorithm for directed graphs.
Furthermore, we leverage structural information from historical state-action trajectories to develop a novel hierarchical decision-making framework based on state and action abstractions applicable to both single-agent and multi-agent scenarios.

\subsection{State Abstraction}
Value function approximation \citep{mahadevan2005value, mahadevan2007proto} utilizes spectral analysis of the state space to derive low-dimensional, compact representations of the Markov Decision Process, thereby improving policy learning. 
The successor representation characterizes states by their discounted occupancy and visitation density, facilitating knowledge transfer across tasks \citep{barreto2017successor} and enabling learning conditioned on specific goals \citep{hoang2021successor}.
However, these methods often suffer from instability and poor generalization in noisy, high-dimensional environments, limiting their effectiveness.

Previous studies have investigated aggregation functions that cluster similar states to reduce task complexity \citep{hutter2016extreme, abel2018state}.
Abstract State Transition Graphs (STGs) \citep{de2018abstract} create compact state representations by identifying structurally similar states through encoded features.
The DSAA method \citep{attali2022discrete} achieves end-to-end state-action abstraction based on successor features and max-entropy regularization. 
Nevertheless, these methods rely on prior knowledge about underlying tasks, such as the distance threshold parameter in the DBSCAN algorithm and the assumed N-simplex distribution over abstract states, which limits their stability and effectiveness across diverse scenarios.
Recent studies \citep{zang2022simsr, zhu2022invariant} have explored different learning objectives to refine state representations from environmental observations.
While these methods yield strong representational power, they often prioritize invariance, potentially discarding essential environmental details—such as context-specific features or dynamic variations—leading to inaccurate characterizations of the original decision process \citep{abel2019state}.

In this work, we leverage similarity-based structural relationships between states to achieve adaptive state abstraction, balancing the removal of irrelevant information with the preservation of critical decision-making features for robust policy learning.

\subsection{Hierarchical Decision Making}

\subsubsection{Skill-based Learning}
Skill-based learning utilizes previously acquired behaviors or skills to facilitate efficient exploration in complex decision-making tasks \citep{merel2018neural, merel2020catch}.
The DHRL method \citep{lee2022dhrl} separates the time horizons of low-level and high-level policies using a graph structure, improving training efficiency in standard environments.
Meanwhile, Gaussian prior \citep{pertsch2021accelerating} and conditional generative models \citep{singh2020parrot} are introduced to approximate the distribution of relevant skills for a given state and to guide exploration of the skill space, respectively.

Despite these benefits, skill discovery remains a fundamental challenge due to the difficulty of identifying reusable and transferable behaviors across diverse tasks.
To address this challenge, researchers have explored structured approaches for constructing meaningful skills.
One approach \citep{jinnai2019discovering} constructs skills by minimizing cover time—a metric derived from the state transition graph—to accelerate learning in environments with sparse rewards.
Reward-respecting subtasks \citep{sutton2023reward} incorporate environmental rewards associated with terminating states, ensuring alignment with the overall reward structure.
The ROD algorithm \citep{machado2023temporal} employs spectral analysis and clustering on successor representations to identify skills that facilitate temporally extended exploration.
However, these methods typically rely on a fixed skill set, limiting adaptability in dynamic environments. 
Furthermore, their dependence on the original state space constrains scalability, particularly in complex environments with high-dimensional and noisy state spaces.

In this work, we define and optimize the structural entropy of directed transitions between abstract states to construct a skill hierarchy with an adaptive scale.
The DSAA algorithm \citep{attali2022discrete} and the Louvain algorithm \citep{evans2023creating} were recently introduced to similarly compute skills between abstract states and to construct multi-level skill hierarchies, respectively.
However, their limited consideration of transition-based relationships between abstract states, along with their reliance on a fixed-resolution partitioning parameter, constrains their effectiveness and flexibility in decision-making across diverse evaluation scenarios, as discussed in Section \ref{subsection: skill-based learning}.

\subsubsection{Role-based Learning}
In natural systems \citep{gordon1996organization, jeanson2005emergence, butler2012condensed}, individuals assume specialized roles that enhance labor division and improve operational or cognitive efficiency. 
Inspired by this, multi-agent systems decompose tasks and assign specialized agents to subtasks, thereby reducing design complexity and improving performance \citep{wooldridge2000gaia, cossentino2014handbook}.

However, the practical implementation of these approaches is often constrained, as predefined task decompositions and roles are not always available in real-world environments, making their application challenging \citep{lhaksmana2018role, sun2020reinforcement}.
To address this, Bayesian inference \citep{wilson2010bayesian} has been incorporated into MARL algorithms to infer roles dynamically, while the ROMA methodology \citep{wang2020roma} fosters role emergence by designing a specialization objective that encourages agents to differentiate their tasks.
Despite these advances, exhaustive exploration of the entire state-action space remains computationally prohibitive. 
The RODE method alleviates this issue by decomposing joint action spaces \citep{wang2020rode}, facilitating more efficient exploration and coordination among agents through a focus on relevant action subspaces.
Nevertheless, its effectiveness is strongly influenced by domain-specific knowledge, as it remains highly sensitive to parameters such as the number of clusters and the choice of distance metrics in clustering algorithms.

In this work, we leverage structural similarities between actions to achieve hierarchical action abstraction, enabling an adaptive, role-based learning method that requires no prior knowledge. 
By dynamically structuring the action space, our approach improves scalability and generalization across diverse multi-agent collaboration scenarios.

\begin{figure}[t]
    \centering
    \includegraphics[width=\textwidth]{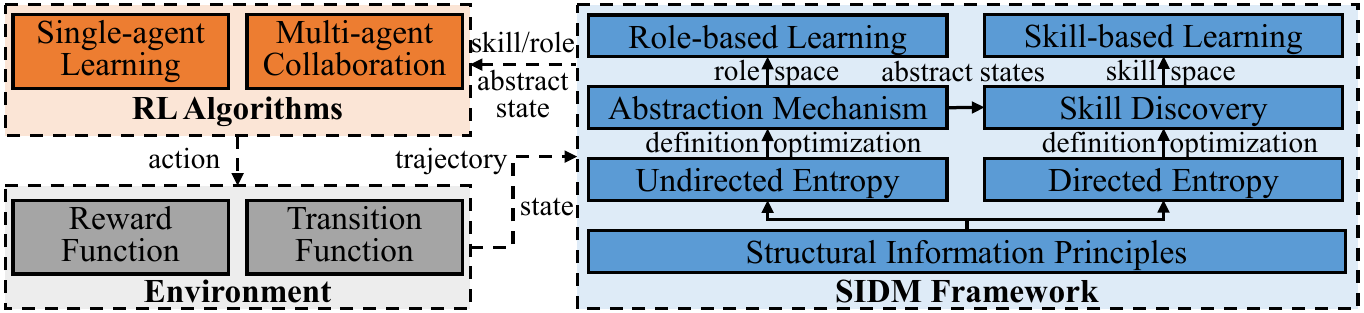}
    \vspace{-0.5cm}
    \caption{The overall decision-making process, including the environment, the proposed SIDM framework, and various downstream RL algorithms.}
    \label{fig: SIDM}
    \vspace{-0.5cm}
\end{figure}

\section{The Proposed SIDM Framework} \label{section: framework}
This section provides an overview of our proposed SIDM framework, which takes as input the original states and historical trajectories from the environment and outputs the discovered skills, roles, and corresponding abstract states, which can be utilized by downstream RL algorithms, as shown in Figure \ref{fig: SIDM}.
Specifically, our framework introduces an adaptive abstraction mechanism to derive abstract representations of states and actions, defines and optimizes directed structural entropy in abstract transitions to facilitate hierarchical skill discovery, and develops skill-based and role-based learning methods for single-agent and multi-agent decision-making, respectively.

\subsection{Abstraction Mechanism} \label{subsection: abstraction mechanism}
To improve learning efficiency and reduce the complexity of high-dimensional and noisy environmental information, we present an adaptive abstraction mechanism based on structural information principles.
This mechanism partitions similar states and actions into distinct communities, producing compact and meaningful abstract state and action representations.

At each timestep $t$, we sample a batch of size $n$ from the replay buffer $\mathcal{B}$, which includes the current state variable $S_t$, action variable $A_t$, subsequent state variable $S_{t+1}$, and reward variable $R_t$.
Each of these denotes a batch of random variables drawn from the replay buffer.
We first employ an encoder-decoder architecture \citep{cho2014learning} to transform the original high-dimensional variables $S_t$, $S_{t+1}$, and $A_t$ into their low-dimensional representations, denoted as $S_t^\prime$, $S_{t+1}^\prime$, and $A_t^\prime$, respectively.
The transformation process is formally defined as follows:
\begin{equation}
    S_t^\prime = f_{\phi_s}^s(S_t)\text{,}\quad S_{t+1}^\prime = f_{\phi_s}^s(S_{t+1})\text{,}\quad A_t^\prime = f_{\phi_a}^a(A_t)\text{,}
\end{equation}
where $f_{\phi_s}^s$ and $f_{\phi_a}^a$ denote the state and action encoders with parameters $\phi_s$ and $\phi_a$, respectively.
To highlight similarities between action functionalities and state transitions, we introduce two decoders, each with a distinct objective: the state decoder $d_s$ predicts actions based on state transitions, while the action decoder $d_a$ reconstructs the next state based on actions and current states.
Specifically, the state decoder $d^s_{\theta_s}(a_t|s^\prime_t,s^\prime_{t+1};\theta_s)$ employs a cross-entropy loss for inverse dynamics modeling \citep{allen2021learning} to predict the current action $a_t \in A_t$ given the adjacent state representations $s_t^\prime \in S_t^\prime$ and $s_{t+1}^\prime \in S_{t+1}^\prime$.
The action decoder $d^a_{\theta_a}(s_{t+1}|s^\prime_t,a_t^\prime;\theta_a)$ reconstructs the next state $s_{t+1} \in S_{t+1}$ using the action representation $a_t^\prime \in A_t^\prime$ and the current state representation $s_t^\prime \in S_t^\prime$.
The decoder loss $\mathcal{L}_{de}$ is computed as follows:
\begin{equation} \label{equ: loss_de}
    \mathcal{L}_{de} = - \frac{1}{n} \cdot \sum_{i=1}^{n} \left[{||d^s_{\theta_s}(s_t^\prime, s_{t+1}^\prime) - a_t||}_2^2 + {||d^a_{\theta_a}(s_t^\prime, a_t^\prime) - s_{t+1}||}_2^2\right]\text{.}
\end{equation}
For simplicity, we denote the combined set of abstract state representations, $S_t^\prime$ and $S_{t+1}^\prime$ as $S$, and refer to $A_t^\prime$ as $A$.
In the following discussion, we use state abstraction as an illustrative example, while applying the same methodology to action abstraction.

Although the bisimulation metric \citep{castro2020scalable} effectively captures behavioral equivalence between states, its computational complexity is prohibitively high. 
Therefore, we second adopt a more computationally efficient similarity metric based on Pearson Correlation Analysis.
For each pair of states $s_i, s_j \in S$ with $i \neq j$, we compute their similarity score $\mathcal{C}(s_i,s_j)$ using Pearson correlation over their feature dimensions:
\begin{equation} \label{equ: pca}
    \mathcal{C}(s_{i}, s_{j})=\frac{\mathbb{E}\left((s_{i}-\mu_{s_{i}})(s_{j}-\mu_{s_{j}})\right)}{\sigma_{s_{i}} \sigma_{s_{j}}}\text{,}
\end{equation}
where $\mu_{s_i}$ and $\mu_{s_j}$ denote the means, and $\sigma_{s_i}$ and $\sigma_{s_j}$ denote the variances of state representations $s_i$ and $s_j$, respectively.
A higher absolute value of $\mathcal{C}(s_i,s_j)$ indicates greater similarity between states $s_i$ and $s_j$, guiding the state abstraction process by grouping similar states together.

Third, we construct a complete, weighted, and undirected state graph $G_s$, where each state $s \in S$ is a vertex and each edge $(s_i, s_j)$ is weighted by the metric $\mathcal{C}(s_i,s_j)$.
To eliminate the interference of insignificant connections, particularly those with absolute values near zero, we apply edge filtration to the complete state graph $G_s$.
Following the prior study \citep{li2016three}, we simplify $G_s$ into a $\text{k}$-nearest neighbor ($\text{k}$NN) graph by minimizing its one-dimensional structural entropy. 
The filtration procedure, summarized in Algorithm \ref{alg: filtration}, involves treating each state $s \in S$ as a center vertex and retaining only its $\text{k}$ edges with the highest absolute weights to construct the $\text{k}$NN graph $G_\text{k}$ (line $4$ in Algorithm \ref{alg: filtration}).
We then compute the one-dimensional structural entropy $H^1(G_\text{k})$ for the resulting graph $G_\text{k}$ (line $5$ in Algorithm \ref{alg: filtration}).
To determine the optimal parameter $\text{k}^*$, we evaluate $H^1(G_\text{k})$ across a range of plausible $\text{k}$ values and select the one that minimizes entropy (lines $3$ and $8$ in Algorithm \ref{alg: filtration}). 
The optimized graph $G_{\text{k}^*}$ serves as the final sparse state graph $G^*_s$ (lines $9$ and $10$ in Algorithm \ref{alg: filtration}).

\begin{algorithm}[t]
    \caption{The Edge Filtration Algorithm}
    \label{alg: filtration}
    \begin{algorithmic}[1]
        \STATE {\bfseries Input:} a weighted, undirected, and complete graph $G=(V,E,W)$
        \STATE {\bfseries Output:} a sparsified graph $G^{*}$
        \FOR{$\text{k}=1$ to $n-1$}
            \IF{the \text{k}NN graph $G_\text{k}$ exists}
                \STATE $H^1(G_\text{k}) \gets$ calculate the one-dimensional structural entropy of $G_\text{k}$ 
            \ENDIF
        \ENDFOR
        \STATE $\text{k}^{*} \gets \arg \min_{k}\{H^1(G_\text{k})\}$
        \STATE $G^{*} \gets G_{\text{k}^*}$
        \RETURN {$G^{*}$}
    \end{algorithmic}
\end{algorithm}

\begin{figure}[t]
    \centering
    \includegraphics[width=\textwidth]{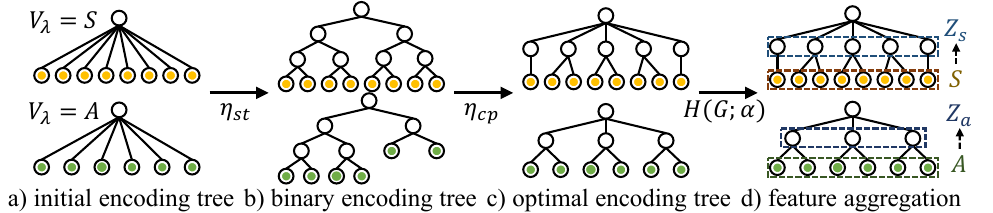}
    \vspace{-0.5cm}
    \caption{The abstraction mechanism for states and actions.}
    \label{fig: abstraction}
    \vspace{-0.5cm}
\end{figure}

Fourth, as illustrated in Figure \ref{fig: abstraction}, we determine the optimal partitioning structure of the sparse state graph $G_s^*$ and design an aggregation function derive abstract states within each community.
We initialize a one-layer partitioning structure for $G_s^*$, represented as the initial encoding tree $T_s$, where each community initially consists of a single state.
To further optimize this partitioning structure, we introduce two stable-enhancing operators from the HSCE algorithm \citep{pan2021information}: \textit{stretch} ($\eta_{st}$) and \textit{compress} ($\eta_{cp}$), which iteratively reduce the structural entropy of $G_s^*$ under $T_s$, increasing the tree height from $1$ to $2$.
The optimization process is summarized in Algorithm \ref{alg: optimization}.
During a ``\textit{stretch}-\textit{compress}" cycle, we operate on the set $U^s_i$ of tree nodes at layer $i$ in $T_{s}$ and quantify the average variation in structural entropy as $\overline{\operatorname{Spar}}_i(T_s)$.
In each iteration, we traverse all sets of tree nodes at the same level and select the set that results in the greatest reduction in structural entropy (line $4$ in Algorithm \ref{alg: optimization}).
These selected nodes then undergo a ``\textit{stretch}-\textit{compress}" cycle (lines $8$ - $13$ in Algorithm \ref{alg: optimization}).
The iteration terminates when either the tree height reaches $h_{T_s}=2$ (line $3$ in Algorithm \ref{alg: optimization}) or no node set exhibits a positive entropy reduction, i.e., $\overline{\operatorname{Spar}}_{i^*}(T_s)>0$ (lines $5$ and $6$ in Algorithm \ref{alg: optimization}).
The process outputs the optimal encoding tree $T_s^{*}$ (lines $19$ and $20$ in Algorithm \ref{alg: optimization}).

\begin{algorithm}[t]
    \caption{The Undirected Optimization Algorithm}
    \label{alg: optimization}
    \begin{algorithmic}[1]
        \STATE {\bfseries Input:} a one-layer initial encoding tree $T$
        \STATE {\bfseries Output:} an optimized two-layer encoding tree $T^*$
        \WHILE{the tree height $h_T < 2$}
            \STATE $i^* \gets \arg \max_{i}\{\overline{\operatorname{Spar}}_i(T)\}$
            \STATE \textbf{\# determine whether the iteration is terminated}
            \IF{$\overline{\operatorname{Spar}}_{i^*}(T) = 0$}
                \STATE break
            \ENDIF
            \STATE \textbf{\# execute optimizations on selected nodes}
            \FOR{$\alpha \in U_{i^*}$}
                \STATE $T \gets \eta_{st}(T_\alpha)$
                \STATE $T \gets \eta_{cp}(T_\alpha)$
                \STATE $h_T \gets h_T + 1$
            \ENDFOR
            \STATE \textbf{\# adjust and update tree structure}
            \FOR{$i = i^* + 1$ to $h_T$}
                \STATE update $U_{i}$
            \ENDFOR
        \ENDWHILE
        \STATE $T^* \gets T$
        \RETURN $T^*$
    \end{algorithmic}
\end{algorithm}

In the optimal encoding tree $T^*_s$, the structural entropy assigned to each node, as specified in Equation \ref{equ: kd_node_se}, quantifies the uncertainty of a single-step random walk reaching its associated vertex community from its parent's community.
Treating this uncertainty as the node weight, we design an aggregation function to compute the node representations in $T^*_s$.
For each leaf node $\nu \in T_s^*$ with a corresponding state set $V_\nu={s}$, its representation is defined as $h_\nu = s$.
For each non-leaf node $\alpha$, we normalize the weights of its children using the softmax function to ensure a probabilistic distribution, then compute its node representation as follows:
\begin{equation} \label{equ: aggregation}
    h_{\alpha}=\sum_{i=1}^{L_\alpha} \frac{\exp\left({H^{T^*_s}\left(G^*_s ; \alpha_i\right)}\right)}{\sum_{j=1}^{L_\alpha}\exp{\left({H^{T^*_s}\left(G^*_s ; \alpha_j\right)}\right)}} \cdot h_{\alpha_i}\text{.} 
\end{equation}
The representation of each child node $h_{\lambda_i}$ of the root node $\lambda$ is defined as an abstract state $z^s_i$, where its corresponding state set is denoted as $S_i \subset S$.
Thus, the set of abstract states is defined as $Z_s=\{z^s_1, z^s_2, \dots, z^s_{L_\lambda} \}$.

At timestep $t$, the original environmental state $s_t \in \mathcal{S}$ is embedded into a low-dimensional vector and mapped to an abstract state $z^s_t$ based on its dot products with the abstract representations in $Z_s$, as follows:
\begin{equation} \label{equ: state abstraction}
    z_t^s = \arg \max_{z^s_i \in Z_s} \langle f_{\phi_s}^s (s_t), z^s_i \rangle \text{,}
\end{equation}
where $f_{\phi_s}^s$ is the encoding function for states.
Similarly, the abstract action variable $Z_a$ is defined as $Z_a = \{z_1^a, z_2^a, \dots, z^a_{L_\lambda} \}$, where each abstract action $z_i^a$ corresponds to a subset $A_i \subset A$. 
In the discrete action space $\mathcal{A}$,  the original actions associated with the embedded actions in $A_i$ collectively form an action subspace, denoted as $\mathcal{A}_i \subset \mathcal{A}$.

\begin{algorithm}[t]
\caption{The Directed Graph Adjustion Algorithm}
\label{alg: directed adjustion}
\begin{algorithmic}[1]
\STATE {\bfseries Input:} a directed, weighted graph with non-negative edge weights $G_{dir} = (V, E_{dir}, W_{dir})$
\STATE {\bfseries Output:} a strongly connected, directed graph $G^\prime_{dir}$ with normalized edge weights
\STATE \textbf{\# identify strongly connected components (SCCs)}
\STATE $SCCs \gets \text{Kosaraju-SCC}(G)$
\STATE $G_{SCC} = (V_{SCC}, E_{SCC}) \gets$ create a meta-graph of connected components
\STATE \textbf{\# ensure strong connectivity between SCCs}
\FOR{$(SCC_i, SCC_j)$ that lack a directed path in $G_{SCC}$}
    \STATE $G_{dir} \gets$ introduce a minimal-weight edge from $SCC_i$ to $SCC_j$
\ENDFOR
\STATE \textbf{\# normalize edge weights to weighted out-degree sum to $1$}
\FOR{$(v_i, v_j) \in E_{dir}$}
        \STATE $W_{dir}(v_i, v_j) \gets W_{dir}(v_i, v_j) / \sum_{(v_i, v_k) \in E} W_{dir}(v_i, v_k)$
    \ENDFOR
\STATE $G_{dir}^\prime \gets G_{dir}$
\RETURN $G_{dir}^\prime$
\end{algorithmic}
\end{algorithm}

\subsection{Directed Structural Entropy} \label{subsection: directed entropy}
To address the limitations posed by undirected constraints in existing structural information principles \citep{li2016structural, zeng2023effective, zeng2023hierarchical}, we define and optimize high-dimensional structural entropy for directed graphs, enabling subsequent accurate representation of key transition dynamics between abstract states in RL problems.

Given a directed graph $G_{dir}=(V, E_{dir}, W_{dir})$ with non-negative edge weights, we introduce Algorithm \ref{alg: directed adjustion} to modify its structure with two primary objectives: (i) to ensure its strong connectivity, meaning there exists a directed path between any pair of vertices, and (ii) to make it more conducive to a random walk process by normalizing edge weights such that the weighted out-degree sum of each vertex is $1$.
To achieve this, we first identify the strongly connected components of $G_{dir}$ (lines $3$ and $4$ in Algorithm \ref{alg: directed adjustion}) and introduce directed edges with minimal weights to establish connectivity between these components (lines $5$ - $9$ in Algorithm \ref{alg: directed adjustion}).
Then, for each vertex $v \in V$, we normalize the weights of all outgoing edges by dividing each weight by the vertex's weighted out-degree sum (lines $11$ - $13$ in Algorithm \ref{alg: directed adjustion}).
The following proposition confirms the existence and uniqueness of the stationary distribution $\pi_s$ over the vertices of the adjusted graph $G_{dir}^\prime$, forming the basis for the definition and optimization of directed structural entropy.
\begin{proposition} \label{theorem: std}
    Given a strongly connected, directed graph $G^\prime_{dir}=(V, E^\prime_{dir}, W^\prime_{dir})$ with non-negative edge weights, where the weighted out-degree of each vertex sums to $1$, its stationary distribution $\pi_s$ exists and is unique.
    This distribution corresponds to the unique eigenvector associated with the dominant eigenvalue $1$ of the adjacency matrix $A^\prime_{dir}$.
\end{proposition}
Two distinct proofs of this proposition, one based on the Perron-Frobenius theorem and the other on the properties of Markov chains, are provided in Appendix \ref{app: std proof}.

For the strongly connected graph $G^\prime_{dir}$, we compute the stationary distribution $\pi_s$ over all vertices and define the one-dimensional directed structural entropy as follows:
\begin{equation}
    H^1(G^\prime_{dir})=-\sum_{v \in V}{\pi_s(v) \cdot \log \pi_s(v)}\text{,}
\end{equation}
where $\pi_s(v)$ denotes the stationary probability of vertex $v$ in $G^\prime_{dir}$.
Using the stationary distribution $\pi_s$, we refine the terms $g_\alpha$ and $\mathcal{V}_\alpha$ for each non-root node $\alpha$ in the encoding tree $T_{dir}$ of $G^\prime_{dir}$, following Equation \ref{equ: kd_node_se}, and redefine its assigned entropy as follows:
\begin{equation} \label{equ: v_term}
    \mathcal{V}_\alpha = \sum_{v_i \in V} \sum_{v_j \in V_\alpha} \left[\pi_s(v_i) \cdot W^\prime_{dir}(v_i,v_j)\right] \text{,}
\end{equation}
\begin{equation} \label{equ: g_term}
    g_\alpha = \sum_{v_i \notin V_\alpha} \sum_{v_j \in V_\alpha} \left[\pi_s(v_i) \cdot W^\prime_{dir}(v_i, v_j)\right] \text{,}
\end{equation}
\begin{equation}
    H^{T_{dir}}(G^\prime_{dir};\alpha)=-\frac{g_\alpha}{\operatorname{vol}(G^\prime_{dir})} \cdot \log_2 \frac{\mathcal{V}_\alpha}{\mathcal{V}_{\alpha^-}} \text{,}
\end{equation}
where the volume of $G^\prime_{dir}$, $\operatorname{vol}(G^\prime_{dir})$, is defined as the sum of in-degrees and out-degrees of all vertices:
\begin{equation}
    \operatorname{vol}(G^\prime_{dir}) = \sum_{v \in V} (d_v^+ + d_v^-)\text{.}
\end{equation}
Thus, the $K$-dimensional directed structural entropy of $G^\prime_{dir}$ is redefined as follows:
\begin{equation}
    H^{T_{dir}}(G^\prime_{dir}) = \sum_{\alpha \in T_{dir}, \alpha \neq \lambda}{H^{T_{dir}}(G^\prime_{dir};\alpha)}\text{,}\quad H^K(G^\prime_{dir}) = \min_{T_{dir}}\left\{H^{T_{dir}}(G^\prime_{dir})\right\}\text{,}
\end{equation}
% \begin{equation}
%     H^K(G^\prime_{dir}) = \min_{T_{dir}}\left\{H^{T_{dir}}(G^\prime_{dir})\right\}\text{.}
% \end{equation}
where $T_{dir}$ ranges over all encoding trees with a maximal height of $K$.

Expanding upon the \textit{merge} ($\eta_{mg}$) and \textit{combine} ($\eta_{cb}$) operators introduced by deDoc \citep{li2018decoding}, we develop an optimization approach for directed structural entropy $H^{T_{dir}}(G^\prime_{dir})$ to determine the optimal tree-structure partitioning strategy $T^*_{dir}$. 
To calculate the entropy variation caused by a single \textit{merge} or \textit{combine} operation on sibling nodes $\alpha, \beta \in T_{dir}$, we introduce $g_{\alpha, \beta}$ defined as the total weight of edges connecting vertices in $V_\alpha$ to vertices in $V_\beta$ as follows:
\begin{equation}
    g_{\alpha,\beta} = \sum_{v_i \in V_\alpha} \sum_{v_j \in V_\beta} \pi_s(v_i) \cdot W^\prime_{dir}(v_i,v_j) \text{.}
\end{equation}
The entropy variation caused by a single \textit{merge} operation on sibling nodes $\alpha, \beta \in T_{dir}$ is denoted as $\Delta_{mg}(T_{dir}, \alpha, \beta)$ and is calculated as follows:
\begin{equation} \label{equ: se_var_merge}
    \Delta_{mg}(T_{dir},\alpha,\beta) = \frac{g_{\alpha, \beta} + g_{\beta, \alpha}}{\operatorname{vol}(G^\prime_{dir})} \cdot \log_2 \frac{\mathcal{V}_{\alpha^-}}{\mathcal{V}_{\mu_{mg}}} - \frac{\sum_{i \neq j}^{L_\alpha} g_{\alpha_i, \alpha_j}}{\operatorname{vol}(G^\prime_{dir})} \cdot \log_2 \frac{\mathcal{V}_{\mu_{mg}}}{\mathcal{V}_{\alpha}} - \frac{\sum_{i \neq j}^{L_\beta} g_{\beta_i, \beta_j}}{\operatorname{vol}(G^\prime_{dir})} \cdot \log_2 \frac{\mathcal{V}_{\mu_{mg}}}{\mathcal{V}_{\beta}}\text{,}
\end{equation}
where $\mu_{mg}$ is the newly added node via the \textit{merge} operation.
The entropy variation caused by a single \textit{combine} on sibling nodes $\alpha, \beta \in T_{dir}$ is denoted as $\Delta_{cb}(T_{dir},\alpha,\beta)$ and is calculated as follows:
\begin{equation} \label{equ: se_var_combine}
    \Delta_{cb}(T_{dir},\alpha,\beta) = \frac{g_{\alpha,\beta} + g_{\beta,\alpha}}{\operatorname{vol}(G^\prime_{dir})} \cdot \log_2 \frac{\mathcal{V}_{\alpha^-}}{\mathcal{V}_{\mu_{cb}}} \text{,}
\end{equation}
where $\mu_{cb}$ is the newly added node via the \textit{combine} operation.
Detailed deviation of these entropy variations is provided in Appendix \ref{app: der_se_var}.

We summarize the iterative optimization process in Algorithm \ref{alg: directed optimization}. 
At each iteration, we traverse all pairs of sibling nodes (lines $5$ and $11$ in Algorithm \ref{alg: directed optimization}) and selectively execute either the \textit{merge} or \textit{combine} operator (lines $7$ and $13$ in Algorithm \ref{alg: directed optimization}), choosing the operation that maximally reduces structural entropy while ensuring the tree height remains below $K$ (line $3$ in Algorithm \ref{alg: directed optimization}). 
When no node pair satisfies $\Delta_{mg}(T_{dir}, \alpha, \beta) > 0$ or $\Delta_{cb}(T_{dir}, \alpha, \beta) > 0$ (lines $6$ and $12$ in Algorithm \ref{alg: directed optimization}), we terminate the iteration and output $T$ as the optimal encoding tree $T^*$ (lines $18$ and $19$ in Algorithm \ref{alg: directed optimization}).

\begin{algorithm}[t]
    \caption{The Directed Optimization Algorithm}
    \label{alg: directed optimization}
    \begin{algorithmic}[1]
        \STATE {\bfseries Input:} an one-layer initial encoding tree $T_{dir}$, $K \in \mathbb{Z}^{+}$
        \STATE {\bfseries Output:} the $K$-layer optimal encoding tree $T_{dir}^*$
        \WHILE{tree height $h_{T_{dir}} < K$}
            \STATE \textbf{\# execute \textit{merge} optimization}
            \STATE $(\alpha^*, \beta^*) \gets \arg \max \{\Delta_{mg}(T_{dir},\alpha,\beta)\}$ via Equation \ref{equ: se_var_merge}
            \IF{$\Delta_{mg}(T_{dir},\alpha^*,\beta^*) > 0$}
                \STATE $T_{dir} \gets \eta_{mg}(T_{dir},\alpha^*,\beta^*)$
                \STATE continue
            \ENDIF
            \STATE \textbf{\# execute \textit{combine} optimization}
            \STATE $(\alpha^*, \beta^*) \gets \arg \max \{\Delta_{cb}(T_{dir},\alpha,\beta)\}$ via Equation \ref{equ: se_var_combine}
            \IF{$\Delta_{cb}(T_{dir},\alpha^*,\beta^*) > 0$}
                \STATE $T_{dir} \gets \eta_{cb}(T_{dir},\alpha^*,\beta^*)$
                \STATE continue
            \ENDIF
            \STATE break
        \ENDWHILE
        \STATE $T_{dir}^{*}\gets T_{dir}$
        \RETURN $T_{dir}^*$
    \end{algorithmic}
\end{algorithm}

Adding directed edges between strongly connected components in the directed graph $G_{dir}$ may alter the original topological relationships, potentially disrupting irreversible environmental constraints among vertices.
To mitigate this issue, Algorithm \ref{alg: directed adjustion} assigns significantly smaller weights to these additional edges compared to the original transition edges.
This guarantees minimal interference with environmental constraints.
Specifically, Algorithm \ref{alg: directed adjustion} carefully ensures the stationary distribution $\pi_s$ exists and remains unique, while Algorithm \ref{alg: directed optimization} employs a greedy strategy prioritizing edges with substantial original weights.
As a result, our approach preserves the structural consistency of the encoding tree before and after directed graph adjustment. 
Consequently, our graph adjustment method establishes an effective and principled framework for computing the structural entropy of directed graphs, preserving essential environmental constraints of the original task. 
This ensures the rationality and consistency of subsequent skill-based and role-based learning processes.

\subsection{Skill Discovery} \label{subsection: skill discovery}
\begin{figure}[t]
    \centering
    \includegraphics[width=\textwidth]{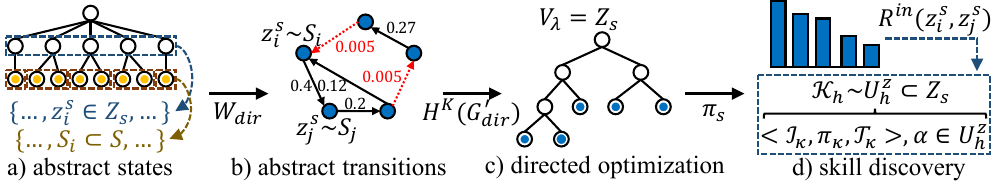}
    \vspace{-0.5cm}
    \caption{The skill discovery based on the directed structural information.}
    \label{fig: skill discovery}
    \vspace{-0.5cm}
\end{figure}

In this subsection, we construct an abstract state transition graph to model the decision-making process and minimize directed structural entropy to obtain the optimal hierarchical partitioning of abstract states.
Within this tree structure, we facilitate skill discovery across hierarchical communities to capture abstract transition dynamics, accounting for the different temporal properties of the RL environment, as illustrated in Figure \ref{fig: skill discovery}.

Taking the abstract states in $Z_s$ as vertices, we construct a weighted, directed graph $G_{dir} = (Z_s, E_{dir}, W_{dir})$, where each vertex represents an abstract state with a corresponding subset of primitive states.
A directed edge $(z_i^s, z_j^s) \in E_{dir}$ is established if there exists an original transition between the corresponding state subsets $S_i$ and $S_j$.
The edge weight $W_{dir}(z_i^s, z_j^s)$ is determined based on the occurrence frequency of these transitions in the sampled batch.
To ensure strong connectivity and normalization, we refine the graph $G_{dir}$ using Algorithm \ref{alg: directed adjustion}, resulting in the adjusted graph $G^\prime_{dir}$.
We apply Algorithm \ref{alg: directed optimization} to derive the optimal encoding tree $T^*_{dir}$, constrained to a maximum height of $K$.

For any integer $0 \leq h \leq K$, representing a hierarchical level in $T^*_{dir}$, the discovered options associated with the node set $U^z_h$ at layer $h$ in $T^*_{dir}$ are defined as follows:
\begin{equation}
    \mathcal{K}_h = \{\langle \mathcal{I}_{\kappa_i}, \pi_{\kappa_i}, \mathcal{T}_{\kappa_i} \rangle \mid \alpha_i \in U^z_h\}\text{.}
\end{equation}
Each option $\kappa_i \in \mathcal{K}_h$ is characterized by its initiation set $\mathcal{I}_{\kappa_i}$, policy function $\pi_{\kappa_i}$, and termination condition $\mathcal{T}_{\kappa_i}$.

In the stationary distribution $\pi_s$, the distribution probability $\pi_s(z^s_j)$ assigned to each abstract state $z^s_j$ reflects its criticality in the historical transition trajectory. 
Abstract states with higher probabilities are more frequently visited and can be considered critical for task completion.
To leverage this notion of state criticality, we define an intrinsic reward function $\mathcal{R}^{in}$ based on the stationary distribution $\pi_s$ over abstract states in $G^\prime_{dir}$.
The reward for transitioning between abstract states $z^s_j$ and $z^s_k$ is computed as follows:
\begin{equation}
    \mathcal{R}^{in}(z^s_j, z^s_k) = \pi_s(z^s_k) - \pi_s(z^s_j)\text{.}
\end{equation}

For each option $\kappa_i \in \mathcal{K}_h$, the termination condition $\mathcal{T}_{\kappa_i}$ consists of abstract states where no further positive intrinsic reward is accumulated, given by:
\begin{equation}
    \mathcal{T}_{\kappa_i}(z_j^s) = 
    \begin{cases}
        1 & \text{if } \arg \max_{z_k^s \in Z^s_{\alpha_i}} \pi_s(z_k^s) = z_j^s\text{,} \\
        0 & \text{otherwise,}
    \end{cases}
\end{equation}
where $Z^s_{\alpha_i} \subset Z_s$ denotes the subset of abstract states corresponding to the node $\alpha_i$ in $T^*_{dir}$.
The initiation set $\mathcal{I}_{\kappa_i}$ is defined as the set of states not included in the termination condition of $\kappa_i$, given by:
\begin{equation}
    \mathcal{I}_{\kappa_i} = \{z_j^s \mid z_j^s \neq \arg \max_{z_k^s \in Z^s_{\alpha_i}} \pi_s(z_k^s)\}\text{.}
\end{equation}
The option policy $\pi_{\kappa_i}$ is trained by maximizing the expected long-term cumulative intrinsic reward, given by:
\begin{equation}
    \pi_{\kappa_i}^* = \arg\max_{\pi_{\kappa_i}} \mathbb{E}_{\mathcal{P}}\left[\sum_{t=0}^{\mathcal{T}_{\kappa_i}(z_j^s)} \gamma^t \mathcal{R}^{in}(z^s_j, z^s_k) \mathcal{T}_{\kappa_i}(z_j^s)\right]\text{,}
\end{equation}
where $z^s_j$ and $z^s_k$ correspond to the abstract states associated with the environmental states $s_t$ and $s_{t+1}$ at timestep $t$, respectively.

Classical skill discovery approaches \citep{jinnai2019discovering, machado2023temporal} employ eigenvalue decomposition to derive the eigenoption $\kappa_e$, which corresponds to the principal eigenvector associated with the largest eigenvalue of the transition matrix $A^\prime_{dir}$ of graph $G^\prime_{dir}$.
The following theorem establishes the relationship between our skill discovery method and these classical approaches to highlight the theoretical properties of our framework.
\begin{theorem} \label{theorem: eigenoption}
    For $h = K$, the discovered set $\mathcal{K}_h$ consists of a single option $\kappa_1$, which is equivalent to the eigenoption $\kappa_e$ associated with the principal eigenvector of the adjusted transition matrix $A^\prime_{dir}$.
\end{theorem}
The detailed proof is provided in Appendix \ref{app: eigenoption proof}.

\begin{figure}[t]
    \centering
    \includegraphics[width=\textwidth]{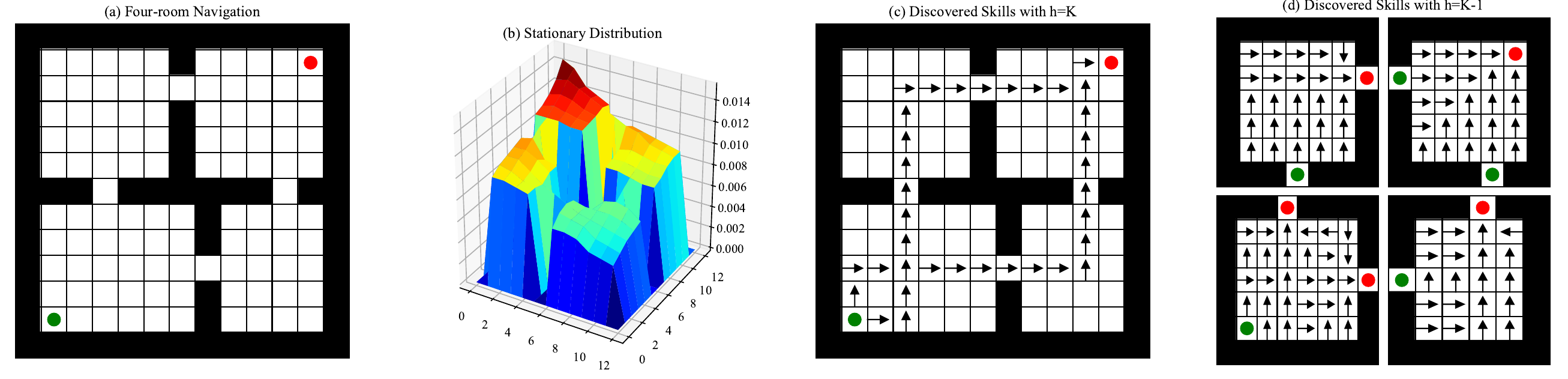}
    \vspace{-0.5cm}
    \caption{The identified skills under different hierarchy levels ($h$ values) in the four-room navigation task.} \label{fig: discovered skills}
    \vspace{-0.3cm}
\end{figure}

By adjusting the parameter $h$, our skill discovery method captures the temporal structure of the environment across different time scales.
Figure \ref{fig: discovered skills} illustrates the discovered options in the four-room domain for varying values of $h$. 
For $h = K$, the sole option in $\mathcal{K}_h$ motivates the agent toward the state with the highest stationary probability in $\pi_s$, which lies along the environment’s diagonal.
For smaller values of $h$, the skill set enables finer-grained navigation over shorter time scales within a subspace of abstract states, where state transitions cover shorter distances, leading to more localized and refined options.

\subsection{Skill-based Learning}
\begin{figure}[t]
    \centering
    \includegraphics[width=\textwidth]{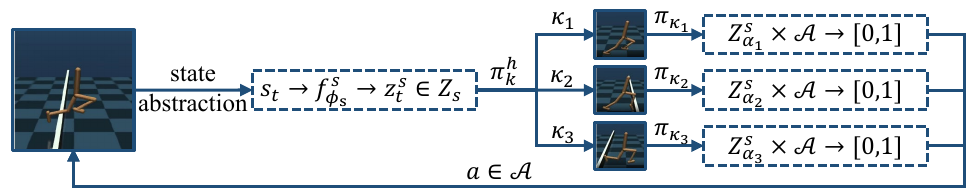}
    \vspace{-0.6cm}
    \caption{The single-agent skill-based learning in SIDM framework.}\label{fig: skill learning}
    %\vspace{-0.2cm}
\end{figure}

In the single-agent decision-making environment, we apply the previously introduced state abstraction and skill discovery mechanisms to construct a hierarchical two-layer skill-based learning framework, as illustrated in Figure \ref{fig: skill learning}.
To model the multi-scale temporal structure of the environment, we define the overall skill set $\mathcal{K}$ as the union of the identified skill sets obtained at hierarchical levels $h = K$ and $h = K - 1$. 
The complete procedure for the proposed two-layer skill-based learning framework is outlined in Algorithm \ref{alg: skill learning}.

\begin{algorithm}[t]
    \caption{The Single-Agent Skill-Based Learning Method}
    \label{alg: skill learning}
    \begin{algorithmic}[1]
        \STATE {\bfseries Input:} batch size $n$, update interval $t_{up}$, replay buffer $\mathcal{B}$
        \STATE {\bfseries Output:} hierarchical skill policies $\pi_\kappa^h$ and $\{\pi_{\kappa_1}, \pi_{\kappa_2}, \dots\}$
        \FOR{each episode}
            \FOR{each timestep $t$}
                \STATE $z_t^s \gets$ obtain an abstract representation of $s_t$ via Equation \ref{equ: state abstraction} 
                \STATE $\kappa_i \gets$ select a skill from the skill space $\mathcal{K}$ based on the high-level policy $\pi_k^h$ 
                \STATE $\tau_t = (s_t, a_t, s_{t+1}, r_t) \gets$ collect transition based on the low-level policy $\pi_{\kappa_i}$ and environmental state $s_t$
                \STATE $\mathcal{B} \gets \mathcal{B} \bigcup \tau_t$
                 \IF{$t \mod t_{up}$ == $0$}
                    \STATE $S_t, A_t, S_{t+1}, R_t \gets$ sample a batch from $\mathcal{B}$
                    \STATE $G_s \gets$ construct the complete state graph from variables $S_t$ and $S_{t+1}$
                    \STATE $G^*_s \gets$ filter out trivial edges from $G_s$
                    \STATE $T^*_s \gets$ generate the optimal encoding tree
                    \STATE $Z_s \gets$ aggregate states to obtain abstract embeddings via Equation \ref{equ: aggregation}
                    \STATE $\mathcal{K} \gets$ update the skill set via the skill discovery method
                    \STATE $\mathcal{L}_{de}, \mathcal{L}_{rl} \gets$ calculate the decoding loss and training loss  
                    \STATE optimize the abstraction function and hierarchical policies by minimizing the loss $\mathcal{L}_{de}$ and $\mathcal{L}_{rl}$ 
                 \ENDIF
            \ENDFOR
        \ENDFOR
    \end{algorithmic}
\end{algorithm}

At each timestep $t$, the agent observes the current environmental state $s_t$ and applies the state abstraction function $f_{\phi_s}^s$ to map $s_t$ to an abstract representation $z_t^s \in Z_s$ (line $5$ in Algorithm \ref{alg: skill learning}).
Based on the abstract state $z_t^s$, the high-level policy $\pi^h_k: Z_s \times \mathcal{K} \rightarrow [0, 1]$ selects an option $\kappa_i$ from the discovered set $\mathcal{K}$ (line $6$ in Algorithm \ref{alg: skill learning}).
The low-level option policy $\pi_{\kappa_i}: Z_{\alpha_i}^s \times \mathcal{A} \rightarrow [0,1]$ maps the abstract state $z_t^s$ to an action distribution over $\mathcal{A}$, from which it samples and executes an action $a_t$.
This results in a transition to a new environmental state $s_{t+1}$ and a received reward $r_t$ (line $7$ in Algorithm \ref{alg: skill learning}).
Low-level policy training is performed using a single-agent underlying RL algorithm that optimizes both environmental and intrinsic rewards.
The associated training loss is denoted as $\mathcal{L}_{rl}$.

At regular update intervals $t_{up}$, we retrieve samples containing state-action transitions $(S_t, A_t, S_{t+1}, R_t)$ of size $n$ from the replay buffer $\mathcal{B}$ (line $10$ in Algorithm \ref{alg: skill learning}).
Next, the state abstraction mechanism (Subsection \ref{subsection: abstraction mechanism}) and skill discovery method (Subsection \ref{subsection: skill discovery}) are used to refine the abstract state set $Z_s$ and update the skill set $\mathcal{K}$ (lines $11$ - $15$ in Algorithm \ref{alg: skill learning}).
Finally, we calculate the training losses $\mathcal{L}_{de}$ and $\mathcal{L}_{rl}$ to optimize the state abstraction function $f_{\phi_s}^s$ and enhance the hierarchical policy learning (lines $16$ and $17$ in Algorithm \ref{alg: skill learning}).

\begin{figure}[t]
    \centering
    \vspace{-0.2cm}
    \includegraphics[width=\textwidth]{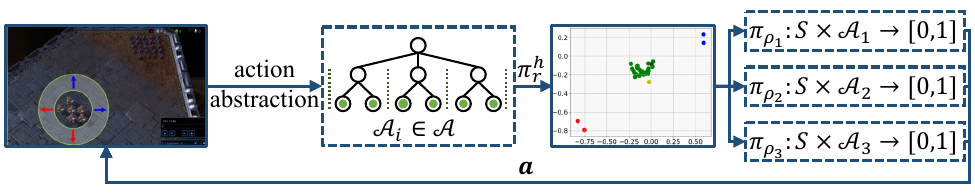}
    \vspace{-0.6cm}
    \caption{The multi-agent role-based learning in SIDM framework.}\label{fig: role learning}
    \vspace{-0.2cm}
\end{figure}

\begin{algorithm}[t]
    \caption{The Multi-Agent Role-Based Learning Method}
    \label{alg: role learning}
    \begin{algorithmic}[1]
        \STATE {\bfseries Input:} batch size $n$, update interval $t_{up}$, replay buffer $\mathcal{B}$
        \STATE {\bfseries Output:} hierarchical role policies $\pi_h^r$ and $\{\pi_{\rho_1}, \pi_{\rho_2}, \dots\}$
        \FOR{each episode}
            \FOR{each timestep $t$}
                \FOR{each agent $n_i \in \mathcal{N}$}
                    \STATE $\rho_j \gets$ select a role from the role space $\Psi$ based on the high-level policy $\pi_r^h$
                    \STATE $a^t_i \gets$ select an action based on the role policy $\pi_{\rho_j}$ and environmental state $s_t$
                \ENDFOR
                \STATE $\mathcal{B} \gets \mathcal{B} \bigcup (s_t, \bm{a}_t, s_{t+1}, r_t)$
                 \IF{$t \mod t_{up}$ == $0$}
                    \STATE $S_t, A_t, S_{t+1}, R_t \gets$ sample a batch from $\mathcal{B}$
                    \STATE $G_a \gets$ construct the complete action graph from the variable $A_t$
                    \STATE $G^*_a \gets$ filter out trivial edges from $G_a$
                    \STATE $T^*_a \gets$ generate the optimal encoding tree
                    \STATE $Z_a \gets$ aggregate actions to obtain abstract embeddings via Equation \ref{equ: aggregation}
                    \STATE $\Psi \gets$ update the role set
                    \STATE $\mathcal{L}_{de}, \mathcal{L}_{marl} \gets$ calculate the decoding loss and training loss
                    \STATE optimize the abstraction function and hierarchical policies by minimizing the losses $\mathcal{L}_{de}$ and $\mathcal{L}_{marl}$ 
                 \ENDIF
            \ENDFOR
        \ENDFOR
    \end{algorithmic}
\end{algorithm}

\subsection{Role-based Learning} \label{subsection: role-based learning}
In fully cooperative multi-agent settings with discrete action spaces, we leverage the state and action abstraction mechanisms described earlier to construct a hierarchical two-layer role-based learning framework, as illustrated in Figure \ref{fig: role learning}.
In this work, we define the set of abstract actions $Z_a$ as the role set $\Psi$, where each role $\rho_i \in \Psi$ corresponds to an action subspace $\mathcal{A}_i \subset \mathcal{A}$. 
The complete procedure for the proposed two-layer role-based learning framework is outlined in Algorithm \ref{alg: role learning}.

At each timestep $t$, for each agent $n_i \in \mathcal{N}$, the high-level policy $\pi^h_r: \tau_i \times \Psi \rightarrow [0,1]$ selects a role $\rho_j$ and its corresponding action subspace $\mathcal{A}_j$ from the role set $\Psi$ to complete the role assignment (line $6$ in Algorithm \ref{alg: role learning}).
The low-level role policy $\pi_{\rho_j}: \mathcal{S} \times \mathcal{A}_j \rightarrow [0,1]$ determines an action distribution over $\mathcal{A}_j$, from which it samples and executes an action $a_i^t \in \mathcal{A}_j$ given the global state $s_t \in \mathcal{S}$ (line $7$ in Algorithm \ref{alg: role learning}).
The individual actions of all agents collectively form a joint action $\bm{a}_t$, which produces a joint reward $r_t$ and leads to the subsequent global state $s_{t+1}$.
To train the low-level role policies, we employ an existing MARL algorithm in which agents sharing the same role utilize a shared policy network. 
The associated training loss is denoted as $\mathcal{L}_{marl}$.

At regular update intervals $t_{up}$, we retrieve samples containing state-action transitions $(S_t, A_t, S_{t+1}, R_t)$ of size $n$ from the replay buffer $\mathcal{B}$ (line $11$ in Algorithm \ref{alg: role learning}).
Subsequently, we apply the action abstraction mechanism to refine the sets $Z_s$ and $\Psi$ (lines $12$ - $16$ in Algorithm \ref{alg: role learning}).
Finally, we calculate the training losses $\mathcal{L}_{de}$ and $\mathcal{L}_{marl}$ to optimize the hierarchical policies (lines $17$ and $18$ in Algorithm \ref{alg: role learning}).

\subsection{Time Complexity Analysis}
This subsection analyzes the time complexity of the abstraction mechanism (Subsection \ref{subsection: abstraction mechanism}) and skill discovery (Subsection \ref{subsection: skill discovery}) in our SIDM framework to evaluate its computational feasibility.
The total time complexity of the abstraction mechanism is given by $O(n^2 + n + m \cdot \log^2 n)$, where $n = |V|$ and $m = |E|$ denote the numbers of vertices and edges, respectively, in the state or action graph.
During algorithm execution, $n$ represents the batch size sampled from the replay buffer, while $m$ denotes the number of edges retained after the sparse operation in Algorithm \ref{alg: filtration}, which is approximately $\frac{\text{k}n}{2}$.
Specifically, graph construction has a complexity of $O(n^{2}+n)$, attributable to $O(n^2)$ for complete graph construction and $O(n)$ for filtering insignificant edges. 
According to the analysis \citep{pan2021information}, the process of optimizing a high-dimensional encoding tree through the \textit{stretch} and \textit{compress} operators incurs a time complexity of $O(m \cdot \log^2 n)$. 
The aggregation of a $K$-layer encoding tree with $n$ leaves has a proven upper bound of $O(n)$. 
Since the number of abstract states in the transition graph does not exceed $n$, the time complexity of skill discovery is capped at $O(m \cdot \log^2 n + n)$.

\section{Experimental Setup} \label{section: setup}
To validate the performance advantage of our SIDM framework, we have conducted extensive comparative experiments in both single-agent decision-making and multi-agent collaboration scenarios. 
Specifically, we evaluate three key components of the SIDM framework: the state abstraction mechanism (SISA), the skill-based learning method (SISL), and the role-based learning method (SIRD).
Each method is compared against state-of-the-art baselines using established and challenging benchmarks.
Each experiment consists of ten independent runs using different random seeds to ensure fairness.
We report the average reward to assess effectiveness and the standard deviation to measure stability over all episodes after convergence.
Additionally, we measure efficiency by counting the environmental timesteps needed to reach specific rewards.

\subsection{Benchmarks}

\subsubsection{State Abstraction}
We evaluate the SISA mechanism for offline state abstraction in a visual Gridworld environment.
Following the approach in \citet{allen2021learning}, each $(x,y)$ coordinate in the $6 \times 6$ Gridworld is linked to a high-dimensional noisy image.
During offline training of the state abstraction mechanism, the agent explores the environment by interacting with these images using a random policy with four directional actions, without access to actual grid positions.
During DQN policy training \citep{mnih2015human}, the abstraction function, which maps original images to abstract state representations, remains fixed.

Following the offline evaluation of the SISA mechanism in Gridworld, we apply it in an online setting to various image-based continuous control tasks from the DeepMind Control Suite (DMControl) \citep{tunyasuvunakool2020dm_control}, where both the abstraction mechanism and policy network are trained simultaneously.
Specifically, our experiments focus on nine DMControl tasks: \texttt{ball\_in\_cup-catch}, \texttt{cartpole-swingup}, \texttt{cheetah-catch}, \texttt{finger-spin}, \texttt{reacher-easy}, \texttt{walker-walk}, \texttt{hopper-hop}, \texttt{hopper-stand}, and \texttt{pendulum-swingup}.

\subsubsection{Skill-based Learning}
We evaluate the skill-based learning method, SISL, in robotic control environments using the MuJoCo physics simulator \citep{todorov2012mujoco}, including a bipedal robot \citep{gehring2021hierarchical} and a 7-DoF fetch arm \citep{silver2018residual}.
For the bipedal robot experiments, we select six tasks requiring diverse skills: \texttt{Hurdles} (jumping), \texttt{Limbo} (torso control), \texttt{Stairs} (intricate foot manipulation), and \texttt{PoleBalance} (body balance).
For the 7-DoF Fetch arm experiments, we select four downstream tasks: \texttt{Table Cleanup}, \texttt{Slippery Push}, \texttt{Pyramid Stack}, and \texttt{Complex Hook}.
To emphasize efficient exploration, all tasks are designed with sparse rewards, which are awarded only when a goal or subgoal is achieved.

\subsubsection{Role-based Learning}
For multi-agent role-based learning, we evaluate the role-based method, SIRD, using the standard Centralized Training with Decentralized Execution (CTDE) benchmark in complex, high-control environments: the StarCraft II micromanagement (SMAC) suite \citep{samvelyan2019starcraft}.
In these micromanagement scenarios, each agent autonomously controls an allied unit based on local observations, while a built-in AI controls all enemy units. 
At each timestep, each agent selects an action from a discrete action space, including four-directional movement, stopping, executing a no-op, and selecting an enemy or ally unit to attack or heal.
The more challenging maps, classified as hard and super-hard, present significant exploration challenges that require intricate collaborative strategies among the agents. 
Thus, we primarily concentrate on the performance of the SIRD method and other baseline approaches in the hard and super-hard maps.
In summary, the multi-agent collaborative SMAC tasks consist of three difficulty levels: easy (\texttt{1c3s5z}, \texttt{2s3z}, \texttt{2c\_vs\_1sc}, and \texttt{10m\_vs\_11m}), hard (\texttt{2c\_vs\_64zg}, \texttt{3s\_vs\_5z}, \texttt{5m\_vs\_6m}, and \texttt{bane\_vs\_bane}), and super hard (\texttt{3s5z\_vs\_3s6z}, \texttt{corridor}, \texttt{MMM2}, and \texttt{27m\_vs\_30m}).

For each benchmark, we provide the task description along with detailed settings for observations, actions, and rewards, as presented in Appendix \ref{app: environment detail}.

\subsection{Baselines}

\subsubsection{State Abstraction}
For offline state abstraction, we select several baseline algorithms from the visual Gridworld environment, each designed to optimize distinct learning objectives. 
These baselines are detailed as follows:

% A model-based RL approach that leverages video prediction to reduce environment interactions while achieving competitive performance in Atari games.
$\bullet$ \textbf{PixelPred \citep{kaiser2019model}}:
A model-based RL approach that leverages video prediction to reduce environment interactions.

% An algorithm that decouples representation learning from policy learning by constructing a compact latent space, improving sample efficiency in image-based RL.
$\bullet$ \textbf{Autoenc \citep{lee2020stochastic}}:
An algorithm that decouples representation learning from policy learning by constructing a compact latent space.

% A state abstraction method that preserves the Markov property using inverse model estimation and temporal contrastive learning, enhancing sample efficiency in RL with rich observations.
$\bullet$ \textbf{Markov \citep{allen2021learning}}:
A state abstraction method that preserves the Markov property using inverse model estimation and temporal contrastive learning.

% A representation model that captures the invariance in action effects across states, improving sample efficiency and generalization by explicitly utilizing action-effect relations.
$\bullet$ \textbf{IAEM \citep{zhu2022invariant}}: 
A representation model that captures the invariance in action effects across states by explicitly utilizing action-effect relations.

For online state abstraction, we adopt state representation and data augmentation algorithms that have demonstrated superior performance in the DMControl Suite, as described in detail below:

% A data augmentation approach that improves sample efficiency and generalization in RL by applying transformations such as random cropping and color jittering to visual inputs.
$\bullet$ \textbf{RAD \citep{laskin2020reinforcement}}: 
A data augmentation approach that applies transformations such as random cropping and color jittering to visual inputs.

% A contrastive learning-based method that extracts high-level features from raw pixels to enhance sample efficiency and control performance in pixel-based RL.
$\bullet$ \textbf{CURL \citep{laskin2020curl}}:
A contrastive learning-based method that extracts high-level features from raw pixels to enhance pixel-based control performance.

% A representation learning technique that utilizes bisimulation metrics to learn task-relevant state representations, improving robustness and generalization in visually complex environments.
$\bullet$ \textbf{DBC \citep{zhang2020learning}}:
A representation learning technique that utilizes bisimulation metrics to learn task-relevant state representations.

% A model-free RL approach that integrates auxiliary autoencoder-based losses to enhance representation learning and stabilize training for high-dimensional image-based tasks.
$\bullet$ \textbf{SAC-AE \citep{yarats2021improving}}:
A model-free RL approach that integrates auxiliary autoencoder-based losses to enhance representation learning.

% A model-free RL algorithm that leverages data augmentation to improve efficiency in visual continuous control, achieving state-of-the-art performance in humanoid locomotion tasks.
$\bullet$ \textbf{DrQv2 \citep{yarats2021mastering}}:
A model-free RL algorithm that leverages data augmentation to improve efficiency in visual continuous control.

% A method that learns robust state representations from image-based observations, showing superior robustness and generalization in visual MuJoCo tasks compared to other solutions.
$\bullet$ \textbf{Simsr \citep{zang2022simsr}}:
A method that learns robust state representations from image-based observations to achieve policy robustness.

% A method that learns disentangled representations by minimizing conditional mutual information between features, improving generalization and training performance in continuous control tasks.
$\bullet$ \textbf{CMID \citep{dunion2024conditional}}:
A method that learns disentangled representations by minimizing conditional mutual information between features.

\subsubsection{Skill-based Learning}
For skill-based learning, we integrate both skill-based and non-skill-based control methods, each demonstrating strong performance in robotic manipulation tasks.
The baselines for the bipedal robot are detailed as follows:

% A model-free deep RL algorithm based on the maximum entropy framework that combines off-policy updates with a stochastic actor-critic formulation to achieve stable, state-of-the-art performance in continuous control tasks.
$\bullet$ \textbf{SAC \citep{haarnoja2018soft}}:
A model-free algorithm using the maximum entropy framework that combines off-policy updates with a stochastic actor-critic formulation.

$\bullet$ \textbf{Switching Ensemble \citep{nachum2019does}}:
A hierarchical RL framework that leverages hierarchy-inspired techniques instead of relying on rigidly imposed structures.

% A hierarchical RL method that leverages off-policy experience for both higher- and lower-level policies, enabling sample-efficient learning of complex tasks with fewer environment interactions.
$\bullet$ \textbf{HIRO \citep{nachum2018data}}:
A hierarchical RL method that leverages off-policy experience for both higher- and lower-level policies.

% A hierarchical RL algorithm that learns task-agnostic options through self-supervised entropy minimization, achieving high sample efficiency and improved success rates on sparse-reward tasks.
$\bullet$ \textbf{HIDIO \citep{zhanghierarchical}}:
A hierarchical RL algorithm that learns task-agnostic options through self-supervised entropy minimization.

% A hierarchical skill learning framework that automatically balances general and specific skills in an unsupervised manner, effectively facilitating skill transfer across a diverse set of tasks.
$\bullet$ \textbf{HSD-3 \citep{gehring2021hierarchical}}:
A hierarchical skill learning framework that automatically balances general and specific skills in an unsupervised manner.

For the 7-DOF robotic arm control task, we select high-performing methods from this benchmark and introduce two additional baselines, DSAA \citep{attali2022discrete} and Louvain \citep{evans2023creating}, that closely resemble our approach. 
This allows for a more direct comparison, further emphasizing the advantages of our skill-based learning method.
These baselines are detailed as follows:

% An RL method that learns a skill prior to offline experience, guiding the exploration of transferable skills in complex navigation and manipulation tasks to improve skill transfer efficiency.
$\bullet$ \textbf{SPiRL \citep{pertsch2021accelerating}}:
An RL method that learns a skill prior to offline experience, guiding the exploration of transferable skills.

% An offline approach that mitigates overestimation bias by combining behavioral cloning with RL, allowing for stable fine-tuning and refined policies without requiring excessive environmental interactions.
$\bullet$ \textbf{BC+Fine-Tuning \citep{beeson2022improving}}:
An offline approach that mitigates overestimation bias by behavioral cloning for stable fine-tuning and refined policies.

% A pre-training method for RL that learns behavioral priors from past tasks, enabling rapid adaptation to new tasks and significantly reducing data requirements in robotic manipulation tasks.
$\bullet$ \textbf{PaRRot \citep{singh2020parrot}}:
A pre-training method for RL that learns behavioral priors from past tasks, enabling rapid adaptation to robotic manipulation tasks.

% A skill-based RL approach that accelerates exploration by using state-conditioned generative models and enables adaptation to unseen task variations using a low-level residual policy.
$\bullet$ \textbf{Reskill \citep{rana2023residual}}:
A skill-based RL approach that accelerates exploration by using state-conditioned generative models and low-level residual policy.

% A method for learning sparse, discrete state-action abstractions through successor representations and max-entropy regularization, which aids in solving tasks with a simpler latent space.
$\bullet$ \textbf{DSAA \citep{attali2022discrete}}:
A method for learning sparse, discrete state-action abstractions through successor representations and max-entropy regularization.

% A hierarchical skill learning method that automatically generates skill hierarchies based on modularity maximization, improving agent learning performance across a wide range of environments.
$\bullet$ \textbf{Louvain \citep{evans2023creating}}:
A hierarchical skill learning method that automatically generates skill hierarchies based on modularity maximization.

For offline pretraining-based baselines (e.g., SPiRL), we follow the warm-start method \citep{zhouefficient}, which consists of pretraining, warmup, and online update phases, to enable efficient fine-tuning of the RL agent without retaining or jointly training on the offline dataset. 
The Louvain baseline, which is limited to discrete state spaces, is integrated into our hierarchical framework by applying Louvain clustering to our similarity-guided state graph to achieve multi-level skill discovery and skill-based learning. 

\subsubsection{Role-based Learning}
For role-based learning, we evaluate both role-based and non-role-based approaches for multi-agent coordination, all of which achieve state-of-the-art performance in the SMAC benchmark.
These baselines are detailed as follows:

% A multi-agent extension of Deep Q-Learning that studies cooperation and competition between autonomous agents learning from raw visual input, particularly in environments like Pong.
$\bullet$ \textbf{IQL \citep{tampuu2017multiagent}}:
A multi-agent extension of Deep Q-Learning that studies cooperation and competition between autonomous agents learning from visual input.

% A cooperative MARL method that decomposes a team’s value function into individual agent values, addressing challenges like spurious rewards and lazy agent behavior in partially-observable environments.
$\bullet$ \textbf{VDN \citep{sunehag2017value}}:
A cooperative MARL method that decomposes a team’s value function into individual agent values, addressing challenges in partially-observable environments.

% A value-based multi-agent method that combines centralized training with decentralized execution by enforcing monotonicity in the joint action values using a mixing network, demonstrating superior performance in the SMAC benchmark.
$\bullet$ \textbf{QMIX \citep{rashid2018qmix}}:
A value-based multi-agent method that combines centralized training with decentralized execution by enforcing monotonicity in the joint action values using a mixing network.

% A MARL method that improves scalability and stability by using a duplex dueling network architecture to efficiently factorize the joint value function while preserving the Individual-Global-Max (IGM) principle.
$\bullet$ \textbf{QPLEX \citep{wang2020qplex}}:
A MARL method that improves scalability and stability by using a duplex dueling network architecture to efficiently factorize the joint value function.

% A MARL approach that generalizes value decomposition by proposing a new factorization method for joint action-value functions, enabling better handling of complex MARL tasks without restrictive structural constraints.
$\bullet$ \textbf{QTRAN \citep{son2019qtran}}:
A MARL approach that generalizes value decomposition by proposing a new factorization method for joint action-value functions without restrictive structural constraints.

% A counterfactual multi-agent actor-critic method that uses a centralized critic and decentralized actors, addressing credit assignment by marginalizing out a single agent’s action while keeping others fixed.
$\bullet$ \textbf{COMA \citep{foerster2018counterfactual}}:
A counterfactual multi-agent actor-critic method that uses a centralized critic and decentralized actors, addressing credit assignment by marginalizing out a single agent’s action while keeping others fixed.

% A MARL algorithm that solves the non-stationarity problem using bidirectional action-dependent Q-learning, turning multi-agent decision-making into a single-agent process, and outperforms state-of-the-art algorithms on benchmarks like SMAC and Google Research Football.
$\bullet$ \textbf{ACE \citep{li2023ace}}:
A MARL algorithm that solves the non-stationarity problem using bidirectional action-dependent Q-learning, turning multi-agent decision-making into a single-agent process.

% A role-based RL method that discovers roles by clustering actions based on their environmental and inter-agent effects, leading to more efficient learning and policy generalization, particularly in the StarCraft II micromanagement benchmark.
$\bullet$ \textbf{RODE \citep{wang2020rode}}:
A role-based RL method that discovers roles by clustering actions based on their environmental and inter-agent effects, leading to more efficient learning and policy generalization.

\subsection{Experimental Setting}
In our proposed SIDM framework, the maximum heights of encoding trees are set to $K=2$ for undirected optimization in state/action abstraction and $K=5$ for directed optimization in skill discovery.

For the SISA mechanism, we set a latent dimension of $2$, a batch size of $2048$, a learning rate of $0.003$, and the Adam optimizer for training the DQN. 
We set a maximum episodic step of $1000$, a batch size of $16$, and a discount factor $\gamma$ for the offline abstraction.  
In the online abstraction, we set a latent dimension of $50$, a replay buffer size of $1e5$, a batch size of $128$, a discount factor of $0.99$, and the Adam optimizer. 
We use the Soft Actor-Critic (SAC) algorithm \citep{haarnoja2018soft} as the underlying single-agent RL method, which is integrated with various state abstraction approaches.  
For the state graph, we set the number of vertices to twice the batch size and determine the number of edges according to Algorithm \ref{alg: filtration}.

For the SISL method, we adopt the standardized SAC algorithm within the corresponding state subspace to train the low-level option policy for each discovered option.  
For the high-level policy, we extract the abstract state with the highest probability within each state subspace to construct the set of termination states. 
We extend the SAC algorithm to the discrete termination set by inputting its continuous output into a Softmax layer. 
The resulting output is a probability distribution over the termination states.
For all experiments, we use neural networks with $4$ hidden layers, skip connections, and ReLU activations.
During the training process, we use the Adam optimizer with a replay buffer size of $1e6$, a mini-batch size of $256$, a learning rate of $0.001$, and a discount factor of $0.99$.  
Similarly, for the state graph, we set the number of vertices to twice the batch size and determine the number of edges according to Algorithm \ref{alg: filtration}.

For the SIRD method, we share a trajectory encoding network with two fully connected layers and a GRU layer for each agent, followed by a linear network without hidden layers or activation functions, which serves as the role policy.  
The outputs of the role policies are fed into separate QMIX-style mixing networks \citep{rashid2020monotonic}, each containing a 32-dimensional hidden layer with ReLU activation, to estimate global action values.  
For all SMAC experiments, the dimension of action representations is set to $20$, the optimizer is set to RMSprop with a learning rate of $0.0005$, and the discount factor is set to $0.99$.  
For the action graph, we set the number of vertices to the number of enemies plus six general discrete actions, and the number of edges is automatically determined according to Algorithm \ref{alg: filtration}.

\subsection{Implementation Details}
We implement the SISA mechanism using Python 3.8.15 and PyTorch 1.13.0, the SISL method using Python 3.9.1 and PyTorch 1.9.0, and the SIRD method using Python 3.5.2 and PyTorch 1.5.1.
All experiments are conducted on five Linux servers, each equipped with an NVIDIA RTX A6000 GPU and an Intel i9-10980XE CPU clocked at 3.00 GHz.

\begin{figure}[t]
    \centering
    \includegraphics[width=1\textwidth]{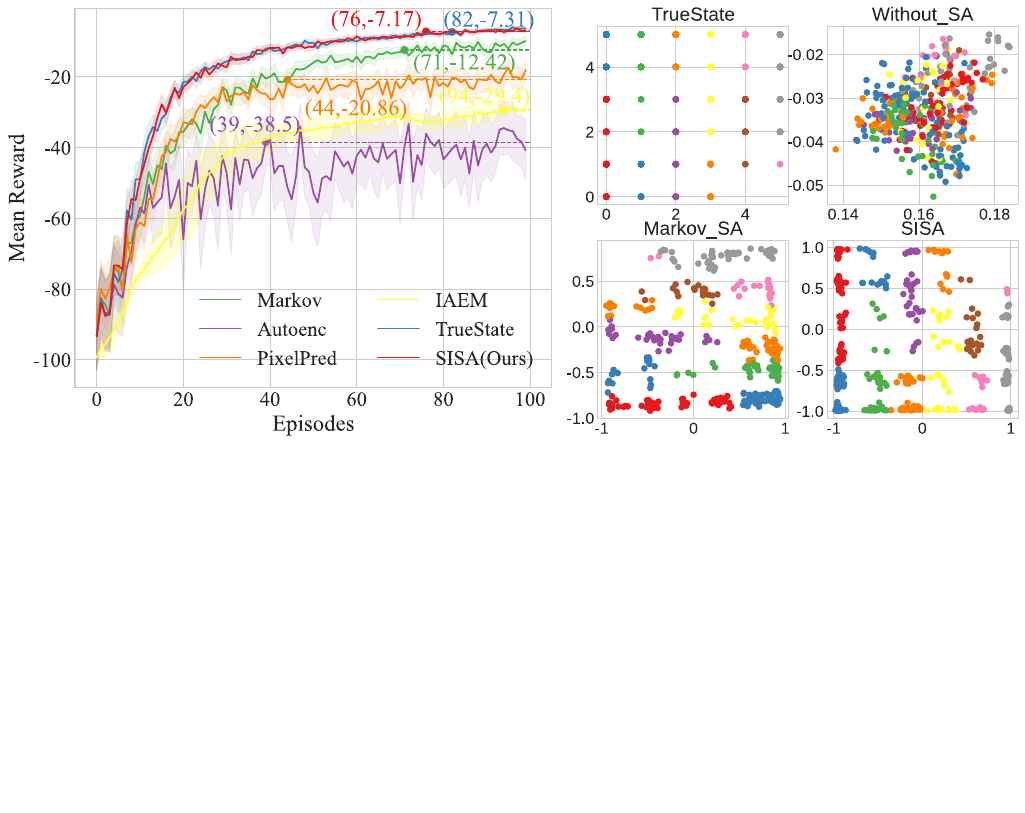}
    \caption{Average rewards over all episodes (left) and visualization of the $2$-dimensional abstract state representations (right) for the navigation task within the $6 \times 6$ Gridworld environment.}
    \label{fig: msa_gridworld}
    \vspace{-0.5cm}
\end{figure}

\section{Evaluation and Discussion} \label{section: evaluation}

\subsection{State Abstraction}
We conduct comparative empirical experiments to demonstrate the benefits of our state abstraction mechanism SISA, focusing on offline abstractions in the visual Gridworld environment and online abstractions in continuous control tasks.

\subsubsection{Offline Abstraction for Visual Gridworld}
In the Gridworld environment, we illustrate the learning curves of SISA and other baselines for the offline navigation task in Figure \ref{fig: msa_gridworld}. 
For reference, we also include the learning curve of the underlying DQN, labeled as TrueState, which is trained on ground-truth positions $(x,y)$ without any observational noise or abstraction function.
For each learning curve, we indicate the convergence point in parentheses within the figure.
As shown in Figure \ref{fig: msa_gridworld} (left), SISA reaches convergence at $76$ epochs with an average episodic reward of -$7.17$, achieving the smallest standard deviation and outperforming all baselines.
Relative to the TrueState performance of $(82,-7.31)$, SISA maintains its advantages in both effectiveness and stability.
Furthermore, Figure \ref{fig: msa_gridworld} (right) presents the $2$-dimensional abstract representations of noisy observations for the $6 \times 6$ Gridworld, with different colors indicating ground-truth positions.
Our abstraction mechanism more accurately reconstructs the relative positions of ground-truth states compared to the baselines.
This success arises from an adaptive trade-off between filtering irrelevant details and preserving essential information, achieved through hierarchical aggregation within the optimal encoding tree.

\begin{table*}[t]
\centering
\caption{Summary of the mean episodic rewards for different tasks from DMControl: ``average value $\pm$ standard deviation" and ``average improvement (absolute value in percentage)". The best performance of our model is highlighted in bold, while the best performance among baselines is underlined.}
\resizebox{1\textwidth}{!}{
\begin{tabular}{c|ccccc}
\hline
Domain, Task & ball\_in\_cup-catch & cartpole-swingup & cheetah-run & finger-spin & reacher-easy \\ \hline
DBC         &     $168.95 \pm 84.76$         &    $317.74 \pm 77.49$     &    $432.24 \pm 181.43$     &    $805.90 \pm 78.85$    &    $191.44 \pm 69.07$ \\
SAC-AE      &     $929.24 \pm 39.14$         &    $\underline{839.23} \pm 15.83$     &    $663.71 \pm 9.16$     &    $898.08 \pm 30.23$    &    $917.24 \pm 38.33$ \\
RAD         &     $\underline{937.97} \pm \underline{6.77}$          &    $825.62 \pm \underline{9.80}$     &    $\underline{802.53} \pm 8.73$     &    $835.20 \pm 93.26$    &    $908.24 \pm \underline{25.62}$ \\
CURL        &     $899.03 \pm 30.61$         &    $824.46 \pm 18.53$     &    $309.49 \pm 8.15$     &    $949.57 \pm 15.71$    &    $\underline{919.71} \pm 28.03$ \\
Markov      &     $919.10 \pm 38.14$         &    $814.94 \pm 17.61$     &    $642.79 \pm 65.92$     &    $\underline{969.91} \pm \underline{8.41}$    &     $806.34 \pm 131.40$ \\
DrQv2       & $275.85 \pm 44.87$ & $573.10 \pm 33.11$ & $583.85 \pm 11.52$ & $629.23 \pm 18.38$ & $406.09 \pm 66.09$ \\ 
Simsr & $837.59 \pm 18.43$ & $803.25 \pm 21.30$ & $708.20 \pm \textbf{2.81}$ & $694.17 \pm 42.84$ & 
 $883.52 \pm 57.19$ \\ 
CMID & $529.03 \pm 37.44$ & $817.49 \pm 14.36$ & $624.78 \pm 15.72$ & $843.74 \pm 21.87$ & $818.49 \pm 33.07$ \\ \hline
SISA$_{pr}$     &  $946.29 \pm 8.63$  &  $858.21 \pm 6.31$  &  $\textbf{806.67} \pm 8.61$  &  $\textbf{970.45} \pm 8.75$  &  $924.52 \pm 19.04$  \\
SISA        &     $\textbf{947.66} \pm \textbf{7.03}$          &    $\textbf{861.37} \pm \textbf{3.26}$      &    $803.32 \pm \underline{5.51}$     &    $968.59 \pm \textbf{6.54}$    &    $\textbf{941.71} \pm \textbf{16.04}$ \\
\hline
Abs.($\%$) Avg. $\uparrow$     &       $9.69(1.03)$        &    $22.14(2.64)$     &     $4.14(0.52)$    &   $0.54(0.06)$     &    $22.0(2.39)$ \\
\hline\cline{1-6}
Domain, Task & walker-walk & hopper-hop & hopper-stand & pendulum-swingup & average reward \\ \hline
DBC         &     $331.97 \pm 108.40$         &    -     &    -     &    $305.08 \pm 86.78$    &    $284.55 \pm 76.43$  \\
SAC-AE      &     $895.33 \pm 56.25$         &    -     &    -     &    -    &    $582.34 \pm 25.07$  \\
RAD         &     $907.08 \pm 13.02$          &    $181.20 \pm \underline{1.80}$     &    $\underline{891.87} \pm \underline{10.04}$     &    $\underline{843.84} \pm 8.99$    &    $\underline{792.61} \pm 19.78$  \\
CURL        &     $885.03 \pm \underline{9.88}$         &    -     &    -     &    -    &    $541.58 \pm \underline{13.69}$ \\
Markov      &     $\underline{918.44} \pm 12.58$        &    $\underline{184.66} \pm 6.48$     &    $864.70 \pm 34.28$     &    $162.58 \pm \underline{1.57}$    &     $698.16 \pm 35.15$  \\
DrQv2       & $588.19 \pm 11.13$ & - & $762.91 \pm 11.31$ & $821.35 \pm 6.60$ & $525.00 \pm 24.10$ \\
Simsr & $804.17 \pm 13.77$ & - & $771.89 \pm 36.27$ & $184.45 \pm 7.23$ & $640.09 \pm 25.46$ \\ 
CMID & $641.83 \pm 27.15$ & - & $173.44 \pm 19.58$ & - & $511.08 \pm 20.77$ \\ \hline
SISA$_{pr}$     &  $\textbf{921.64} \pm 12.43$  &  $209.55 \pm 6.46$  &  $893.54 \pm \textbf{4.74}$  &  $839.19 \pm 7.90$  &  $818.90 \pm 9.21$  \\
SISA        &     $919.78 \pm \textbf{9.40}$          &    $\textbf{209.68} \pm \textbf{6.23}$      &    $\textbf{900.45} \pm 5.05$     &    $\textbf{851.94} \pm \textbf{3.60}$    &    $\textbf{822.72} \pm \textbf{6.96}$ \\
\hline
Abs.($\%$) Avg. $\uparrow$     &       $3.20(0.35)$         &    $25.02(13.55)$     &     $8.58(0.96)$    &   $8.1(0.96)$     &    $30.11(3.80)$ \\
\hline
\end{tabular}}
\label{tab: sisa}
\vspace{-0.5cm}
\end{table*}

\subsubsection{Online Abstraction for Continuous Control}
In the online abstraction experiments, we evaluate SISA and baseline methods across nine continuous tasks from the DMControl suite.
Table \ref{tab: sisa} summarizes the average value and standard deviation of episodic rewards, excluding results with final rewards below $100.00$.
Our results show that SISA consistently outperforms the baselines across all DMControl tasks, achieving up to a $25.02$ increase in mean episodic reward.
This corresponds to a $13.55\%$ improvement from $184.66$ to $209.68$ in the \texttt{hopper-hop} task.
Moreover, SISA exhibits greater stability than other methods, with reduced standard deviations in six tasks.
In the remaining tasks, SISA ranks among the lowest deviations, closely matching the top-performing baselines.
Compared to our previous version, SISA$_{pr}$ \citep{zeng2023hierarchical}, the current method achieves higher average rewards in six tasks and lower standard deviations in eight tasks.
These results demonstrate that the undirected structural entropy optimization algorithm (Subsection \ref{subsection: abstraction mechanism}) enhances both the effectiveness and stability of state abstraction.

\begin{figure}[t]
    \centering
    \includegraphics[width=1\textwidth]{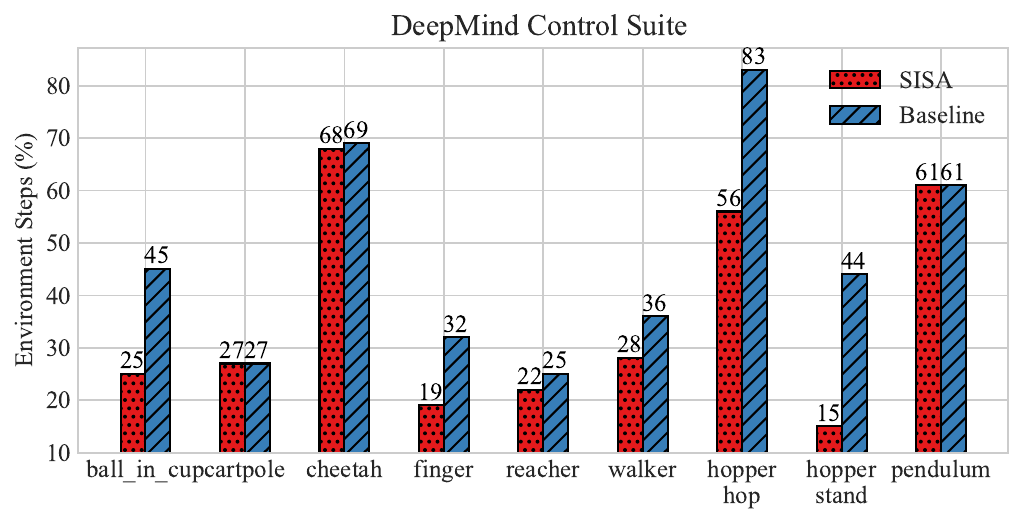}
    \vspace{-1cm}
    \caption{Sample-efficiency analysis of the SISA mechanism and baseline method in the DMControl continuous control tasks.}
    \label{fig: sisa sample efficiency}
    \vspace{-0.5cm}
\end{figure}

To further analyze sample efficiency in the DMControl experiments, we define the target reward as $90\%$ of SISA's final average value and report the timesteps required for both SISA and the best-performing baseline to reach this target.
As shown in Figure \ref{fig: sisa sample efficiency}, SISA reaches the mean episodic reward target in fewer steps than the baseline, demonstrating superior sample efficiency.
Specifically, in the \texttt{hopper-stand} task, SISA improves sample efficiency by $64.86\%$, reducing the required timesteps from $222k$ to $78k$ to reach an episodic reward of $810.41$.
In summary, SISA have established a new state-of-the-art on DMControl, excelling in policy quality, stability, and sample efficiency in online learning with reward-based feedback.
This success stems from our state abstraction mechanism, which optimally balances the compression of irrelevant information with the retention of essential features, enhancing both learning efficiency and policy performance.

\begin{figure*}[t]
    \centering
    \includegraphics[width=1\textwidth]{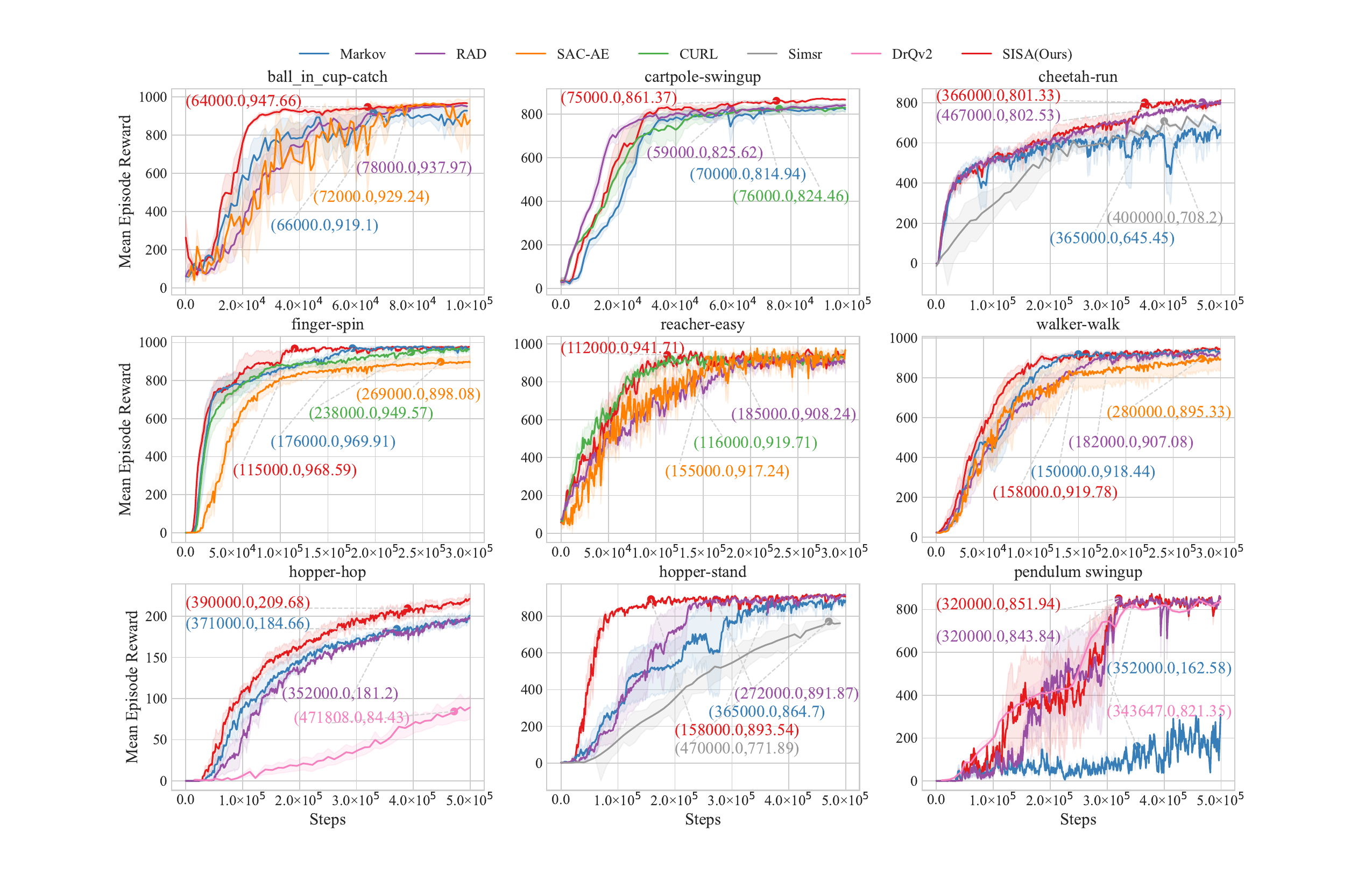}
    \vspace{-0.8cm}
    \caption{Learning curves of SISA and three leading baselines in various DMControl tasks.}
    \label{fig: msa_dmcontrol}
    \vspace{-0.2cm}
\end{figure*}

For each DMControl task, Figure \ref{fig: msa_dmcontrol} presents detailed learning curves for SISA and three leading baselines, showing the progression of mean episodic rewards and their convergence points throughout training.
For example, in the \texttt{pendulum-swingup} task, SISA converges within $320,000$ timesteps and attains a mean reward of $851.94$.

\subsection{Skill-based Learning} \label{subsection: skill-based learning}
\begin{figure*}[t]
    \centering
    \includegraphics[width=1\textwidth]{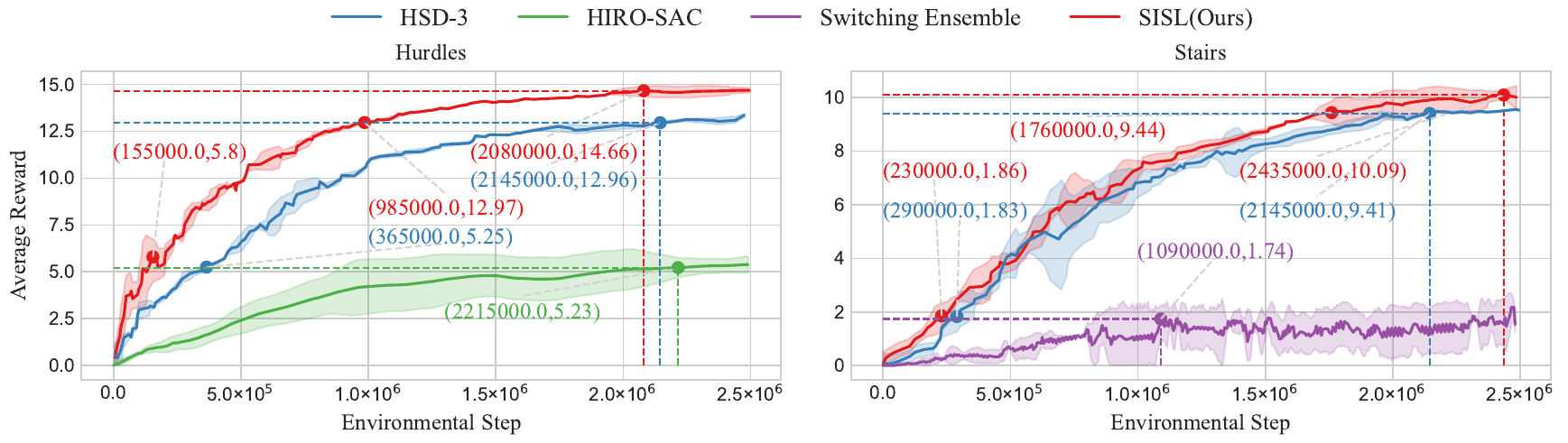}
    \caption{Efficiency comparison of our SISL and baseline methods in the bipedal robotic environment.}
    \label{fig: sisl efficiency}
\end{figure*}

In the bipedal robotic environment, we present the average rewards and standard deviations of SISL and other baselines after one million steps in Table \ref{tab: sisl bpd}.
Notably, SISL consistently outperforms all baseline methods in each control task, achieving a maximum improvement of $18.75\%$ in average reward, increasing from $11.2$ to $13.3$ in the \texttt{Hurdles} task.
To further analyze training efficiency, we visualize the reward learning curves of SISL and the two best-performing baselines in the \texttt{Hurdles} and \texttt{Stairs} tasks.
As shown in Figure \ref{fig: sisl efficiency}, SISL consistently achieves a policy of comparable quality to all baselines while requiring fewer environment steps during training on both tasks.
For instance, in the \texttt{Hurdles} task, to reach a reward of $12.96$, the HSD-3 baseline requires $21.45$ million environmental steps, while SISL achieves the same reward with only $985,000$ steps, highlighting the efficiency advantage of our approach.

\begin{figure*}[t]
    \centering
    % \vspace{-0.2cm}
    \includegraphics[width=1\textwidth]{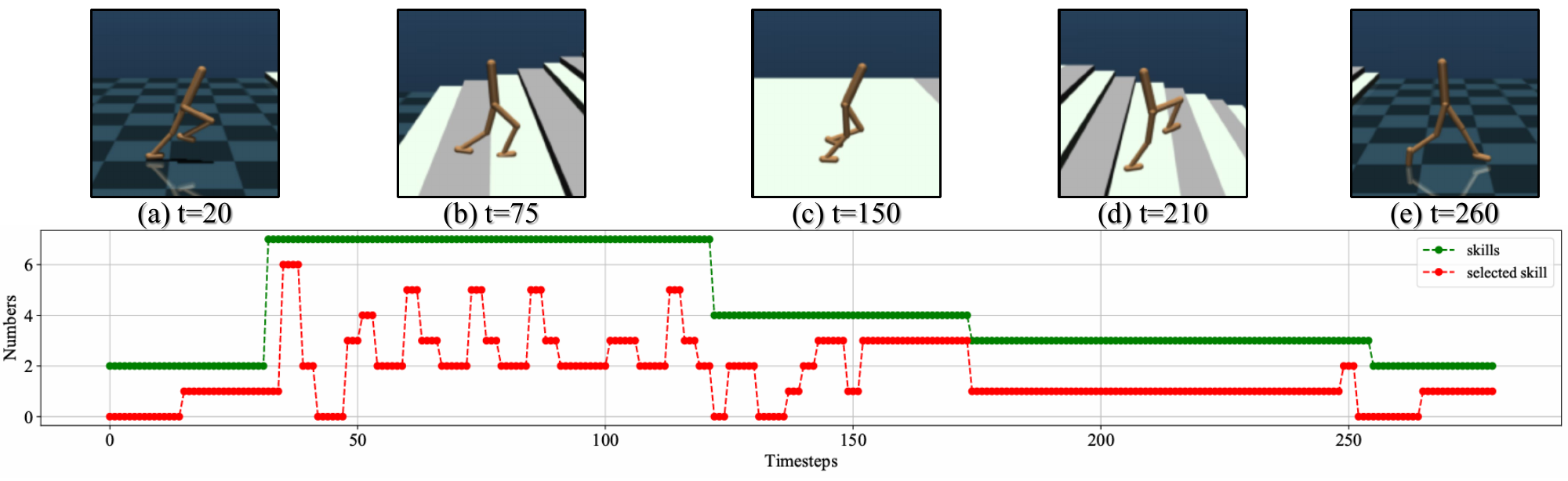}
    \vspace{-0.5cm}
    \caption{The skill discovery and selection during a testing episode of the \texttt{Stairs} task in our single-agent skill-based learning method.}
    \vspace{-0.5cm}
    \label{fig: sisl selection}
\end{figure*}

Moreover, Figure \ref{fig: sisl selection} illustrates the skill selection process in SISL during a testing episode of the \texttt{Stairs} task, highlighting skill discovery dynamics and selection over time.
During the episode, SISL discovers and selects skills with distinct temporal properties across different task phases.
Specifically, walking upstairs requires regular control of the torso’s X and Z positions, with occasional adjustments to the Y position and foot movements.
Running forward follows a different pattern, primarily relying on the X position.
To maintain balance while descending, the right foot is explicitly adjusted more frequently.
SISL dynamically adapts skill discovery and selection to different environmental phases by minimizing structural entropy in directed abstract transitions, successfully completing the primary task without task-specific prior knowledge.
A video demonstration of SISL across multiple episodes and tasks is available on GitHub\footnote{\url{https://selgroup.github.io/SIDM/}}.

\begin{table*}[t]
\centering
\caption{Summary of the final performance across benchmark tasks using the bipedal robot: ``average value $\pm$ standard deviation" and ``average improvement (absolute value in percentage)". The best performance in each category is highlighted in bold, while the second-best performance is underlined.}
\resizebox{1\textwidth}{!}{
\begin{tabular}{c|cccccc}
\hline
Benchmark Task & Hurdles & Limbo & HurdlesLimbo & Stairs & Gaps & PoleBalance \\ \hline
SAC         &     $-0.1 \pm 6.2$         &    $-0.1 \pm \underline{0.2}$     &    $-0.1 \pm 0.4$     &    $0.0 \pm 4.8$    &    $-0.1 \pm 0.5$  &  $231.5 \pm 104.5$ \\
Switching Ensemble  &  $-0.2 \pm 3.0$  &  $-0.2 \pm 4.3$  &  $-0.2 \pm 3.6$  &  $1.1 \pm 3.8$  &  $-0.2 \pm 0.3$  &  $132.8 \pm 230.1$  \\
HIRO-SAC      &     $3.9 \pm 1.6$         &    $1.1 \pm 2.2$     &    $3.5 \pm \underline{0.1}$     &    $0.0 \pm \underline{0.0}$    &    $\underline{0.0} \pm \underline{0.2}$  &  $96.4 \pm \underline{12.4}$ \\
HIDIO        &     $-0.1 \pm \textbf{0.1}$          &    $-0.1 \pm \textbf{0.1}$     &    $-0.2 \pm 0.1$     &    $-0.2 \pm 0.3$    &    $-0.2 \pm 0.3$  &  $117.6 \pm 33.8$ \\
HSD-3   &  $\underline{11.2} \pm 2.0$  &  $\underline{12.0} \pm 0.9$  &  $\underline{11.2} \pm 1.3$  &  $\underline{6.5} \pm 0.7$  &  $-0.2 \pm 8.9$  &  $\underline{246.0} \pm 36.9$ \\
\hline
SISL      &  $\textbf{13.3} \pm \underline{0.9}$  & $\textbf{12.6} \pm 0.7$ &  $\textbf{12.8} \pm \textbf{0.1}$  &  $\textbf{7.0} \pm \underline{0.1}$  &   $\textbf{0.0} \pm \textbf{0.0}$  &  $\textbf{252.6} \pm \textbf{11.2}$ \\
\hline
Abs.($\%$) Avg. $\uparrow$   &  $2.1(18.75)$  &  $0.6(5.0)$  &  $1.6(14.29)$  &   $0.5(7.69)$ &  $0.0(0.0)$  &  $6.6(2.68)$ \\
\hline
\end{tabular}}
\vspace{-0.3cm}
\label{tab: sisl bpd}
\end{table*}

For the $7$-DoF fetching benchmark, we compare SISL with baselines operating in either the original action space (PARROT and BC+FineTuning) or the skill space (SPiRL, Reskill, DSAA, and Louvain).
Table \ref{tab: sisl dof} summarizes the average rewards and standard deviations after convergence across four control tasks, excluding final rewards below $10.00$.
SISL significantly outperforms all baselines in every benchmark task, achieving a maximum improvement of $32.70\%$ from $42.42$ to $56.29$ in the \texttt{Slippery Push} task.
In terms of learning stability, SISL exhibits minimal deviations in the \texttt{Pyramid Stack} and \texttt{Complex Hook} tasks and low deviations in \texttt{Table Cleanup} and \texttt{Complex Hook}, closely trailing the top-performing baseline, Reskill.
Compared with the baseline DSAA, which also leverages abstract states for skill discovery, our SISL method demonstrates improved learning effectiveness, achieving an average improvement of $22.72$ in average reward. 
This is because the skill hierarchy constructed by SISL captures more fine-grained temporal dynamics, leading to better agent exploration and policy optimization. 
Compared to Louvain, which is based on multi-level skill hierarchies, SISL achieves greater learning stability, reducing the standard deviation of final rewards by an average of $68.83\%$, from $3.16$ to $0.99$.
This improvement stems from SISL’s ability to adjust its skill hierarchy while mitigating observational noise through state abstraction, resulting in more stable learning trajectories and reduced performance fluctuations.
Thus, while existing research shares technical similarities with this work in skill discovery via state abstraction or hierarchical skill construction, it further supports the validity of our approach.
Moreover, guided by the structural information principle, we propose a unified hierarchical learning framework that spans from state abstraction to adaptive skill hierarchy construction, while effectively mitigating reliance on prior knowledge and reducing intrinsic observational noise in downstream tasks.

\begin{figure*}[t]
    \centering
    \includegraphics[width=1\textwidth]{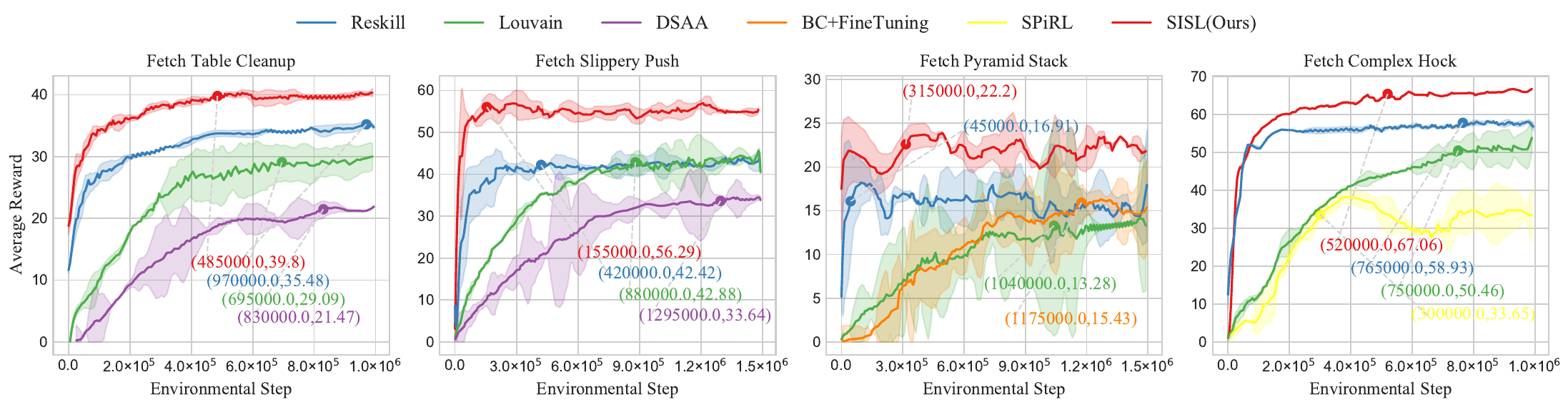}
    \caption{Learning curves of SISL and three leading baselines in various fetching tasks with the 7-DOF robotic arms.}
    \label{fig: arm sisl}
    \vspace{-0.5cm}
\end{figure*}

Figure \ref{fig: arm sisl} presents detailed training learning curves for SISL and three leading baselines across each control task, showing the progression of mean episodic rewards and their convergence points.
For example, in the \texttt{Table Cleanup} task, SISL converges within $485,000$ timesteps and attains a mean reward of $39.80$.

\begin{table*}[t]
\centering
\caption{Summary of the final performances across benchmark tasks with the $7$-DOF robotic arms: ``average value $\pm$ standard deviation" and ``average improvement (absolute value in percentage)". The best performance in each category is highlighted in bold, while the second-best performance is underlined.}
\resizebox{1\textwidth}{!}{
\begin{tabular}{c|cccc}
\hline
Benchmark Task & Fetch Table Cleanup & Fetch Slippery Push & Fetch Pyramid Stack & Fetch Complex Hock \\ \hline
SPiRL & $11.33 \pm 1.74$ & - & - & $33.65 \pm 3.34$ \\
BC + Fine-Tuning & - & $14.81 \pm 2.91$ & $15.43 \pm \underline{2.03}$ & - \\
PARROT & - & $32.10 \pm 3.58$ & - & $13.07 \pm 2.89$ \\
Reskill & $\underline{35.48} \pm \textbf{0.39}$ & $42.42 \pm \textbf{1.32}$ & $\underline{16.91} \pm \underline{2.78}$ & $\underline{58.93} \pm \underline{0.34}$ \\
DSAA & $21.47 \pm 0.38$ & $33.64 \pm 1.72$ & $10.58 \pm 1.97$ & $28.79 \pm 3.02$ \\
Louvain & $29.09 \pm 2.48$ & $\underline{42.88} \pm 2.62$ & $13.28 \pm 3.62$ & $50.46 \pm 3.92$ \\
\hline
SISL & $\textbf{39.80} \pm \underline{0.71}$ & $\textbf{56.29} \pm \underline{1.43}$ & $\textbf{22.20} \pm \textbf{1.67}$ & $\textbf{67.06} \pm \textbf{0.13}$ \\
\hline
Abs.($\%$) Avg. $\uparrow$ & $4.32(12.18)$ & $13.41(32.27)$ & $5.29(31.28)$ & $8.13(13.80)$ \\
\hline
\end{tabular}}
\label{tab: sisl dof}
\end{table*}

\subsection{Role-based Learning}
In this subsection, we compare the SIRD method with state-of-the-art MARL algorithms across SMAC maps in the easy, hard, and super-hard categories. 
Table \ref{tab: sird twr} summarizes the average win rates (above $10.00$) and their standard deviations for each map category.
SIRD outperforms all baseline algorithms, achieving up to a $5.19\%$ improvement in average reward and a reduction of up to $88.26\%$ in standard deviation, highlighting its performance advantages in both learning effectiveness and stability. 
By leveraging structural information principles, our action abstraction mechanism facilitates automatic role discovery, improving agent cooperation and reducing reliance on sensitive hyperparameters.
These improvements are particularly evident in challenging exploration scenarios, such as the hard and super-hard maps.

\begin{table}[t]
\centering
\caption{Summary of the test win rates under different map categories: ``average value $\pm$ standard deviation" and ``improvements/reductions (absolute value in percentage)".  The best performance in each category is highlighted in bold, while the second-best performance is underlined.}
\resizebox{0.75\textwidth}{!}{
\begin{tabular}{c|ccc}
    \hline
    Categories & Easy & Hard & Super Hard \\
    \hline
    COMA & $16.67\pm22.73$ & - & - \\
    IQL & $52.50\pm40.69$ & $73.44\pm24.85$ & $10.55\pm18.49$ \\
    VDN & $85.01\pm17.22$ & $71.49\pm18.78$ & $71.10\pm27.23$ \\
    QMIX & $\underline{98.44}$ $\pm$ $\underline{2.10}$ & $87.11\pm18.58$ & $70.31\pm38.65$\\
    QTRAN & $64.69\pm36.79$ & $58.20\pm45.37$ & $16.80\pm20.61$ \\
    QPLEX & $96.88\pm5.04$ & $89.85$ $\pm$ $\underline{11.35}$ & $84.77\pm10.76$\\
    MAPPO & $66.67\pm35.35$ & $61.72$ $\pm$ $23.60$ & $73.98\pm16.45$\\
    RODE & $93.47\pm10.19$ & $88.44\pm20.96$ & $\underline{92.71}$ $\pm$ $\underline{9.20}$ \\
    ACE & - & $\underline{91.10} \pm 11.66$  & $88.15 \pm 5.21$ \\
    \hline
    SIRD$_{pr}$ &  $98.61\pm\textbf{1.75}$  &  $95.31 \pm 6.63$  &  $95.71\pm3.10$\\
    SIRD & $\textbf{98.83}\pm2.17$ & $\textbf{95.83}\pm\textbf{4.99}$ & $\textbf{97.50}\pm\textbf{1.08}$\\
    \hline
    Abs.($\%$) Avg. $\uparrow$ & $0.39(0.40)$ & $4.73(5.19)$ & $4.79(5.17)$ \\
    Abs.($\%$) Dev. $\downarrow$ & $0.35(16.67)$ & $6.36(56.04)$ & $8.12(88.26)$ \\
    \hline
\end{tabular}}
\label{tab: sird twr}
\end{table}

\begin{figure}[t]
    \centering
    \includegraphics[width=0.9\textwidth]{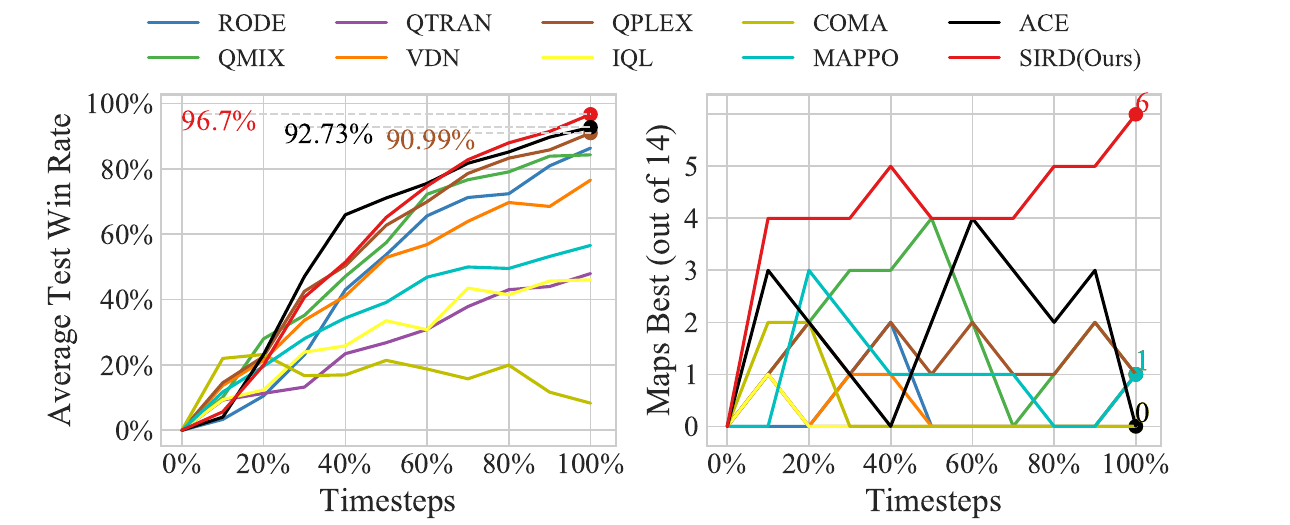}
    \caption{(left) The average test win rates across all 14 maps; (right) The number of maps (out of 14) where the algorithm achieves the highest average test win rate.}
    \label{fig: awr_nbm}
    \vspace{-0.5cm}
\end{figure}

Furthermore, we compare the SIRD method with baseline algorithms across all $14$ SMAC maps to evaluate their overall performance.
Figure \ref{fig: awr_nbm} presents the average test win rate and the number of maps where each MARL algorithm leads in performance at different stages of policy learning.
As shown in Figure \ref{fig: awr_nbm} (left), SIRD surpasses all baselines and achieves faster convergence.
Remarkably, SIRD maintains the highest average test win rate throughout the last $60\%$ of the learning process, achieving a final win rate of $96.7\%$, which exceeds the second-best (ACE at $92.73\%$) and the third-best (QPLEX at $90.99\%$) by $3.97\%$ and $5.71\%$, respectively.
This enhanced performance, particularly in policy quality and learning efficiency, stems from SIRD’s effective exploration of action subsets identified via its action abstraction mechanism.
Figure \ref{fig: awr_nbm} (right) shows that SIRD achieves the best final policy in nearly half the maps ($6$ out of $14$), significantly outperforming the baselines.

\begin{figure*}[t]
    \centering
    % \vspace{-0.2cm}
    \includegraphics[width=1\textwidth]{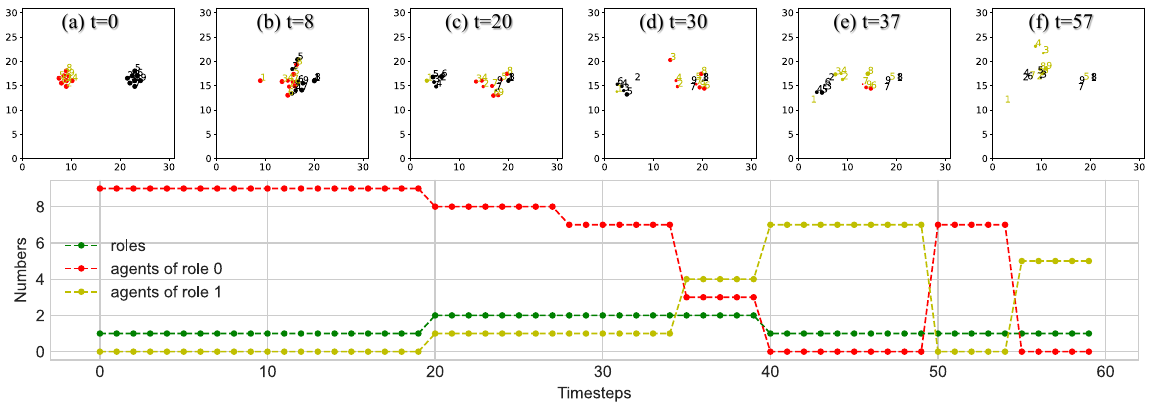}
    \caption{The role discovery and selection during a testing episode of the $1c3s5z$ map in our multi-agent role-based learning method.}
    \vspace{-0.8cm}
    \label{fig: sird collaboration}
\end{figure*}

Figure \ref{fig: sird collaboration} illustrates SIRD-driven multi-agent collaboration in the \texttt{1c3s5z} task, highlighting role variation and agent distribution throughout a testing episode.
Compared to the role-based baseline RODE, SIRD dynamically adjusts its role set and action subspaces via action abstraction, without manual intervention, leading to improved performance.
In summary, SIRD outperforms all MARL baselines in learning effectiveness and stability, establishing a new state-of-the-art on SMAC.

\begin{figure*}[t]
    \centering
    \includegraphics[width=1\textwidth]{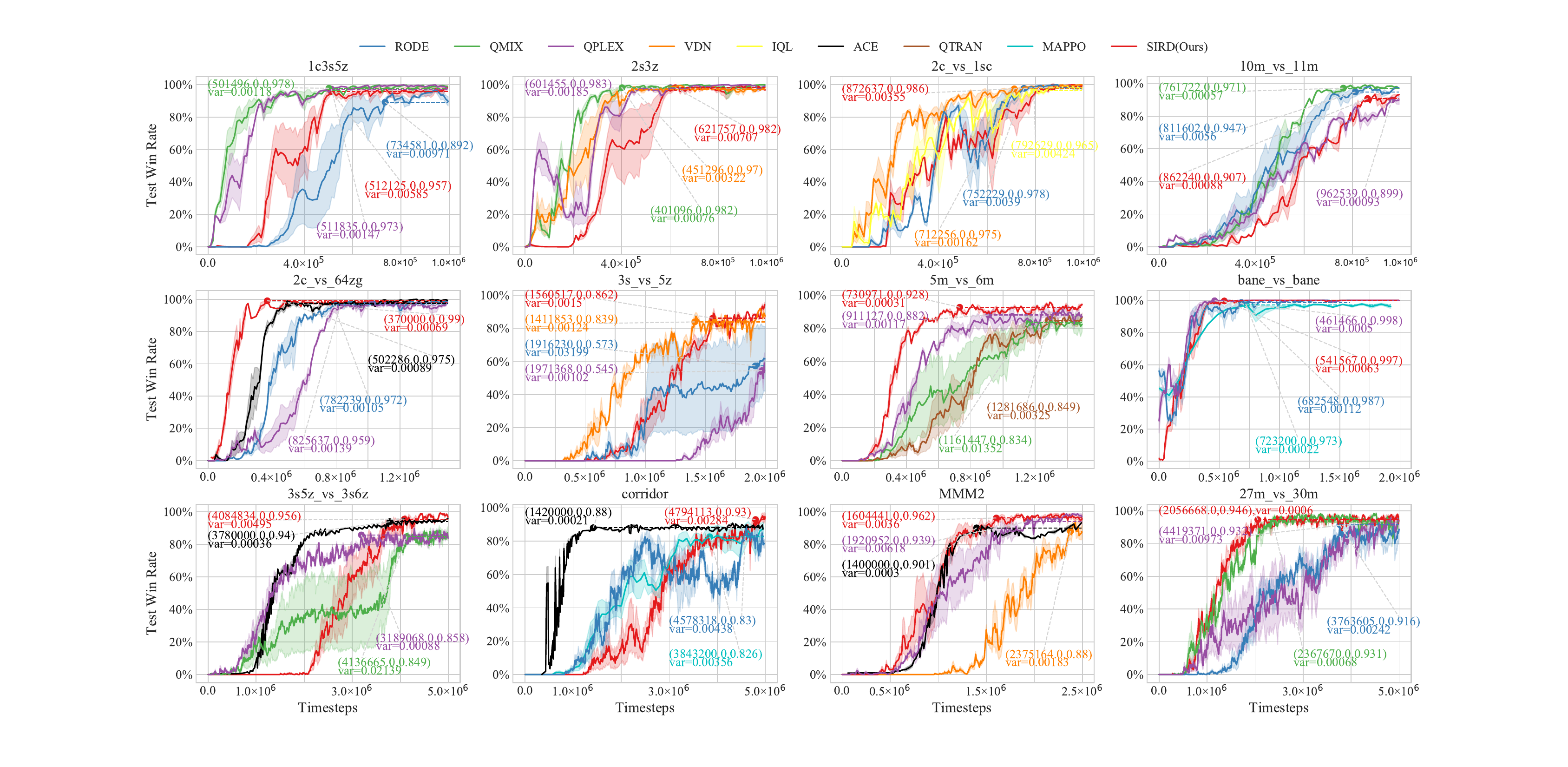}
    \vspace{-0.6cm}
    \caption{Learning curves of SIRD and three leading baselines in various SMAC maps.}
    \vspace{-0.5cm}
    \label{fig: marl}
\end{figure*}

Figure \ref{fig: marl} presents the training curves for SIRD and three leading baselines on each SMAC map, highlighting convergence points and showing standard deviations for each task.
In the \texttt{MMM2} task, SIRD converges after $1,604,442$ timesteps and achieves an average win rate of $96.2\%$.

\subsection{Generality Abilities}
The proposed SIDM is a general framework and can be flexibly integrated with various single-agent and multi-agent RL algorithms, serving as a high-level method for skill and role discovery to enhance hierarchical learning performance.

\begin{figure}[t]
    \centering
    \subfigure[Fetch Slippery Push]{
        \centering
        \includegraphics[width=0.9\textwidth]{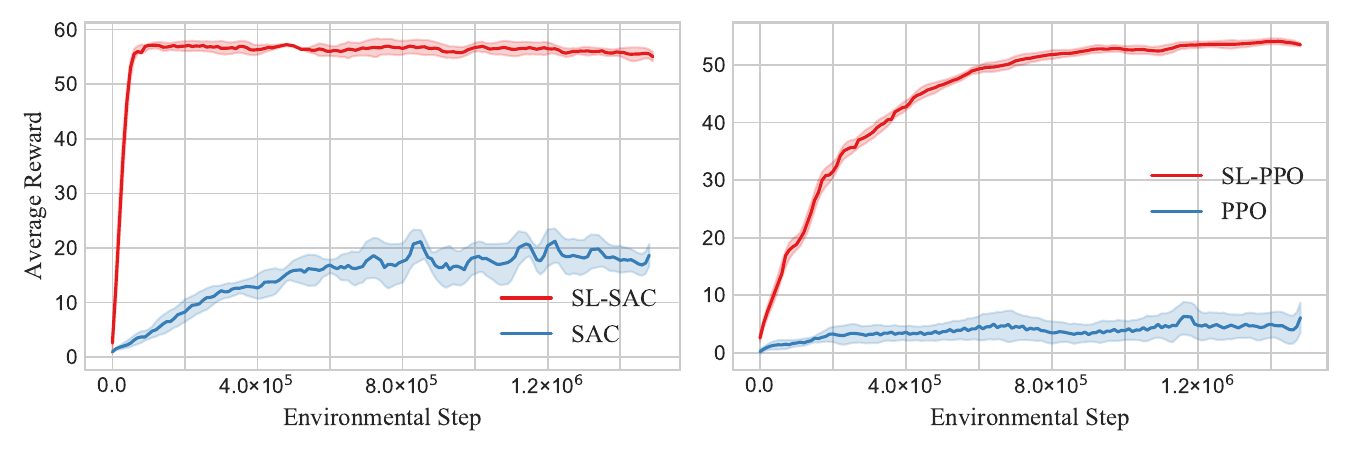}
    }
    \subfigure[Fetch Pyramid Stack]{
        \centering
        \includegraphics[width=0.9\textwidth]{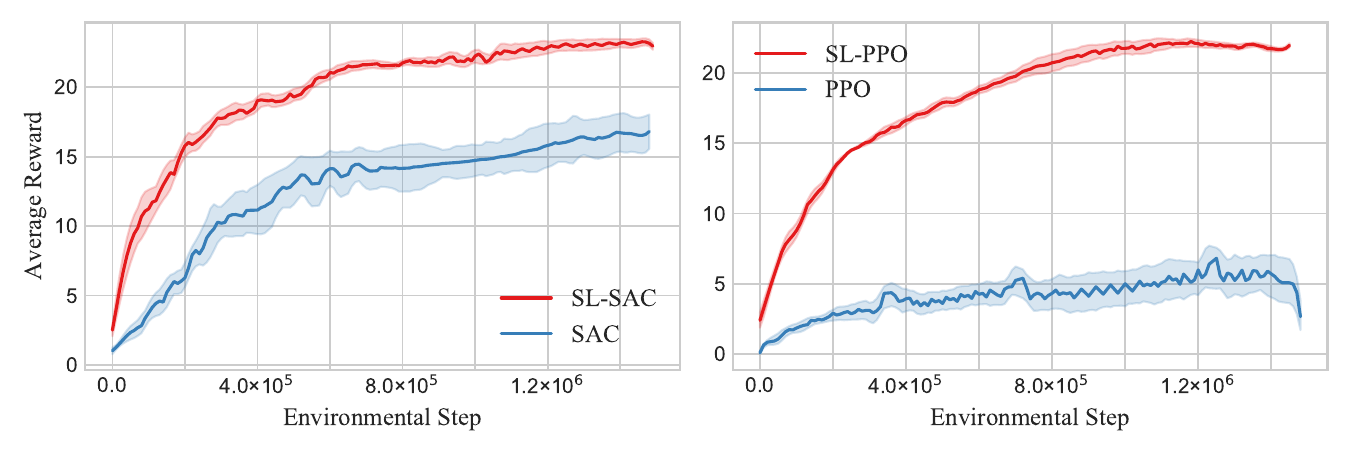}
    }
    \vspace{-0.3cm}
    \caption{Training curves of SISL integrated with single-agent RL methods, specifically SAC and PPO, in fetching tasks with the 7-DOF robotic arms.}
    \label{fig: sisl_integrate}
\end{figure}

\begin{figure}[t]
    \centering
    \subfigure[2c\_vs\_64zg]{
        \centering
        \includegraphics[width=0.9\textwidth]{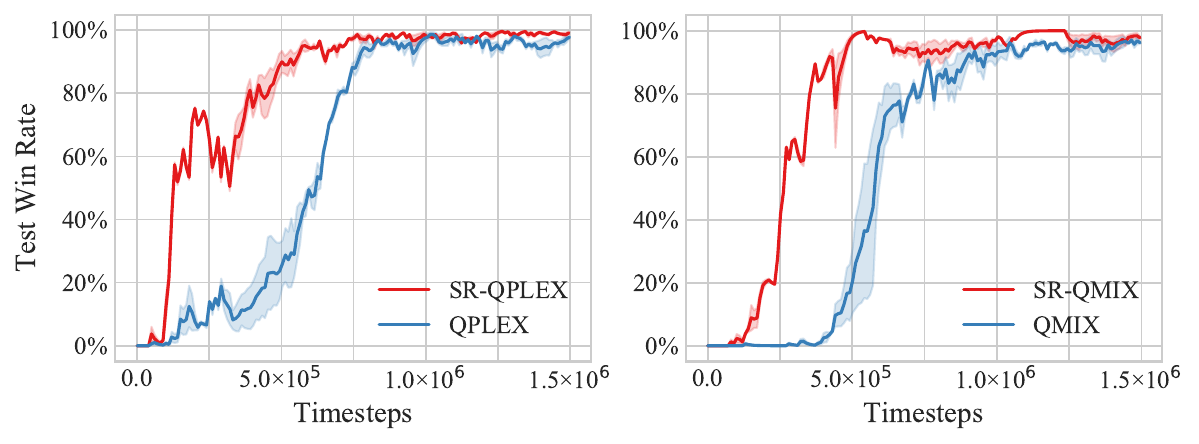}
    }
    \subfigure[MMM2]{
        \centering
        \includegraphics[width=0.9\textwidth]{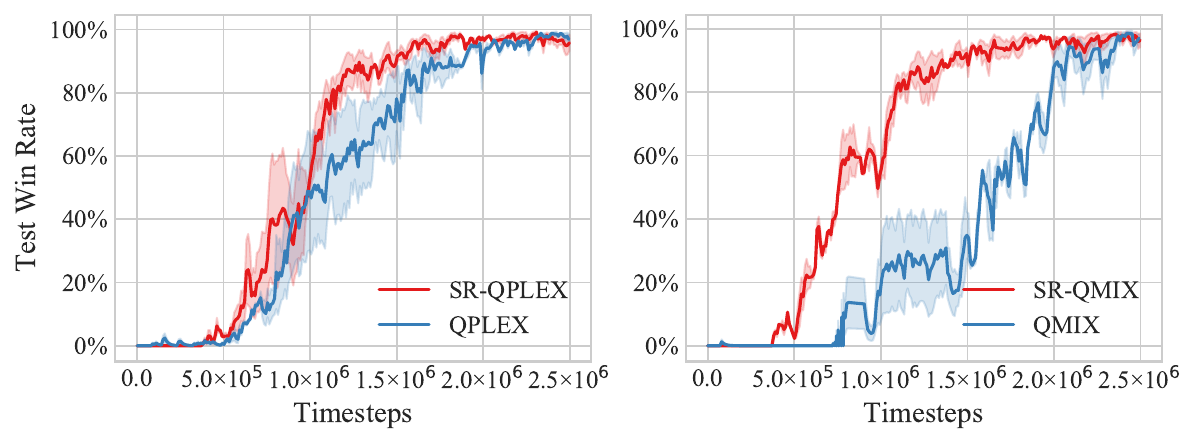}
    }
    \caption{Training curves of SIRD integrated with multi-agent algorithms, specifically QMIX and QPLEX, in SMAC maps.}
    \label{fig: marl_integrate}
\end{figure}

For single-agent decision-making, we employ SAC \citep{haarnoja2018soft} and PPO \citep{schulman2017proximal} as low-level RL algorithms in skill-based learning, forming the SL-SAC and SL-PPO variants.
As shown in Figure \ref{fig: sisl_integrate}, the SL-SAC and SL-PPO variants significantly outperform their original counterparts in terms of learning performance, particularly during training on the \texttt{Slippery Push} and \texttt{Pyramid Stack} tasks in the $7$-DoF Fetching benchmark.
For multi-agent coordination, we integrate the SIRD method with QMIX \citep{rashid2018qmix} and QPLEX \citep{wang2020qplex} algorithms, resulting in the SI-QMIX and SI-QPLEX variants.
As shown in Figure \ref{fig: marl_integrate}, SI-QMIX and SI-QPLEX outperform their original counterparts in terms of policy quality and sample efficiency, particularly on the \texttt{2c\_vs\_64zg} and \texttt{MMM2} maps in the SMAC benchmark.
This leads to faster convergence and more effective multi-agent coordination during training.

These experimental results highlight the effectiveness of applying structural information principles to adaptively identify inherent hierarchical decision-making structures, forming the foundation for general skill-based and role-based learning methods.

\subsection{Ablation Studies}
\begin{figure}[t]
    \centering
    \includegraphics[width=0.9\textwidth]{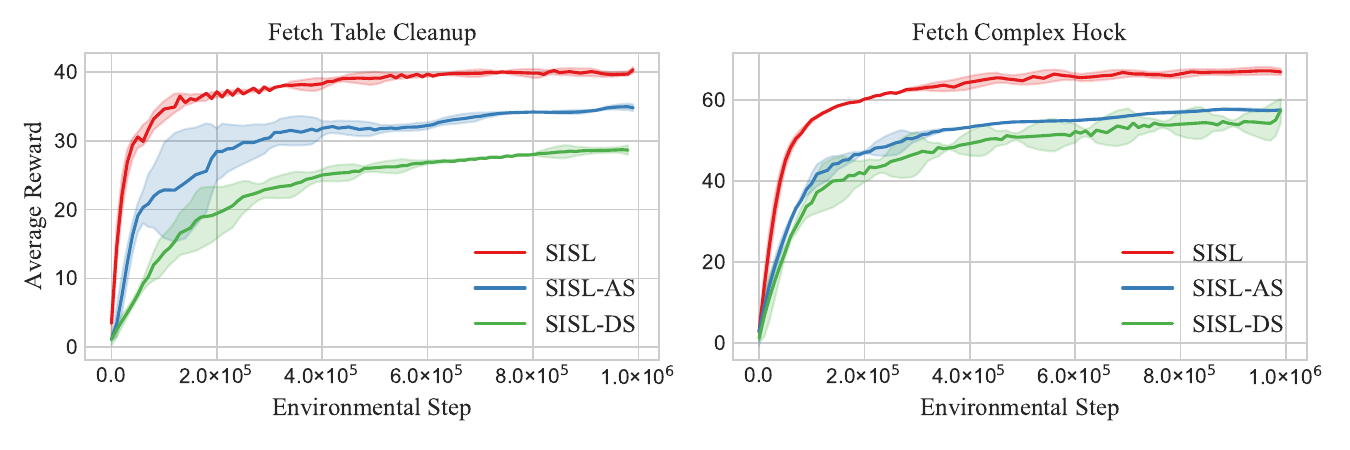}
    \caption{Ablation studies on state abstraction and directed optimization in single-agent, skill-based learning.}
    \label{fig: sisl_ablation}
    \vspace{-0.5cm}
\end{figure}

We evaluate the contribution of state abstraction (Subsection \ref{subsection: abstraction mechanism}) and directed entropy optimization (Subsection \ref{subsection: directed entropy}) to the performance advantages of SIDM in single-agent skill-based learning.
Specifically, we introduce two SISL variants—SISL-AS and SISL-DS—by disabling these components, respectively.
In SISL-AS, the skill set is directly extracted from demonstration data and remains fixed throughout the learning process.
In SISL-DS, the parameter $h$ is set to $K$, and the set $\mathcal{K}_h$ is restricted to a single skill defined at the largest time scale.
This configuration restricts the skill set to a single skill, which is insufficient for learning complex decision-making tasks.
As shown in Figure \ref{fig: sisl_ablation}, the significantly lower training performance of SISL-DS underscores the importance of directed structural entropy in capturing key transition patterns and constructing a skill hierarchy.
The performance gap between SISL and SISL-AS highlights the benefits of adaptive skill discovery and observational noise mitigation through the state abstraction mechanism, as discussed in Subsection \ref{subsection: skill-based learning}.

\begin{figure}[t]
    \centering
    \includegraphics[width=0.9\textwidth]{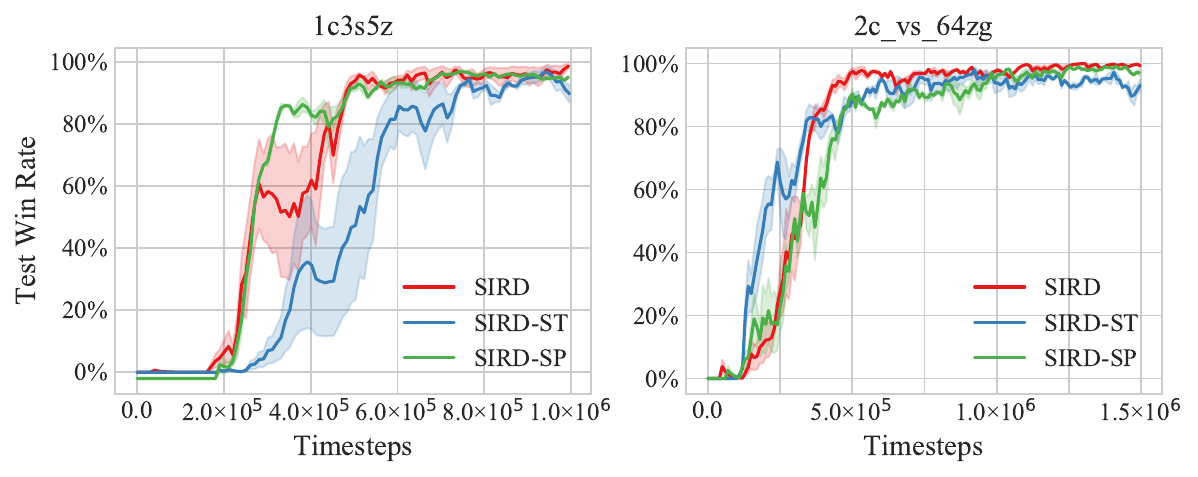}
    \vspace{-0.2cm}
    \caption{Ablation studies on graph construction and edge filtration in multi-agent, role-based learning.}
    \label{fig: marl_ablation}
    \vspace{-0.3cm}
\end{figure}

In multi-agent scenarios, we conduct ablation studies to validate the impact of the graph construction and edge filtration modules in role-based learning (Subsection \ref{subsection: role-based learning}) on the \texttt{2c\_vs\_64zg} and \texttt{1c3s5z} SMAC maps. 
We develop two simplified SIRD variants, SIRD-ST and SIRD-SP, by removing either the graph construction or edge filtration module to isolate their individual contributions.
Specifically, SIRD-ST identifies roles by applying K-means clustering to the joint action space. 
In contrast, SIRD-SP directly optimizes the encoding tree of the complete action graph for role discovery.
As shown in Figure \ref{fig: marl_ablation}, SIRD significantly outperforms SIRD-ST in both mean test win rate and policy stability, emphasizing the crucial role of graph construction in hierarchical learning.
The comparison between SIRD and SIRD-SP suggests that edge filtration accelerates learning by reducing convergence timesteps without compromising performance.

\begin{table*}[t]
\centering
\caption{Sensitivity analysis on alternative similarity metrics in the SISA method using the DMControl suite.}
\begin{tabular}{ccccc}
\hline
Task & \multicolumn{1}{|c}{cheetah-run} & \multicolumn{1}{|c}{finger-spin} & \multicolumn{1}{|c}{reacher-easy} & \multicolumn{1}{|c}{walker-walk} \\ \hline
SISA$_\text{bm}$ & \multicolumn{1}{|c}{$802.94 \pm 7.28$} & \multicolumn{1}{|c}{$968.96 \pm 8.03$} & \multicolumn{1}{|c}{$943.06 \pm 15.82$} & \multicolumn{1}{|c}{$921.61 \pm 11.66$} \\
SISA$_\text{sfs}$ & \multicolumn{1}{|c}{$802.77 \pm 8.04$} & \multicolumn{1}{|c}{$968.14 \pm 8.19$} & \multicolumn{1}{|c}{$938.29 \pm 13.80$} & \multicolumn{1}{|c}{$920.71 \pm 10.53$} \\
SISA$_\text{hrd}$ & \multicolumn{1}{|c}{$803.19 \pm 6.91$} & \multicolumn{1}{|c}{$970.03 \pm 10.58$} & \multicolumn{1}{|c}{$939.52 \pm 18.29$} & \multicolumn{1}{|c}{$923.08 \pm 14.47$} \\
SISA$_\text{crs}$ & \multicolumn{1}{|c}{$802.59 \pm 7.40$} & \multicolumn{1}{|c}{$970.49 \pm 8.95$} & \multicolumn{1}{|c}{$941.33 \pm 14.09$} & \multicolumn{1}{|c}{$923.55 \pm 13.70$} \\ \hline
SISA & \multicolumn{1}{|c}{$803.32 \pm 5.51$} & \multicolumn{1}{|c}{$968.59 \pm 6.54$} & \multicolumn{1}{|c}{$941.71 \pm 16.04$} & \multicolumn{1}{|c}{$919.78 \pm 9.40$} \\ \hline
\end{tabular}
\vspace{-0.2cm}
\label{tab: similarity}
\end{table*}

\begin{table*}[t]
\centering
\caption{Sensitivity analysis on encoding tree height in SIRD method using the SMAC benchmark.}
\resizebox{\textwidth}{!}{
\begin{tabular}{ccccccc}
\hline
\multirow{2}{*}{Category, Map} & \multicolumn{2}{|c}{Easy Maps} & \multicolumn{2}{|c}{Hard Maps} & \multicolumn{2}{|c}{Super Hard Maps} \\ \cline{2-7}
& \multicolumn{1}{|c}{1c3s5z} & \multicolumn{1}{c}{2s3z} & \multicolumn{1}{|c}{2c\_vs\_64zg} & \multicolumn{1}{c}{3s\_vs\_5z} & \multicolumn{1}{|c}{MMM2} & \multicolumn{1}{c}{27m\_vs\_30m} \\ \hline
SIRD-2 & \multicolumn{1}{|c}{$95.72 \pm 0.59$} & \multicolumn{1}{c}{$98.24 \pm 0.71$} & \multicolumn{1}{|c}{$99.04 \pm 0.07$} & \multicolumn{1}{c}{$86.18 \pm 0.15$} & \multicolumn{1}{|c}{$96.20 \pm 0.36$} & \multicolumn{1}{c}{$94.61 \pm 0.06$} \\ \hline
SIRD-3 & \multicolumn{1}{|c}{$95.07 \pm 0.76$} & \multicolumn{1}{c}{$97.60 \pm 0.77$} & \multicolumn{1}{|c}{$96.78 \pm 0.25$} & \multicolumn{1}{c}{$84.53 \pm 0.41$} & \multicolumn{1}{|c}{$96.83 \pm 0.19$} & \multicolumn{1}{c}{$95.47 \pm 0.13$} \\ \hline
SIRD-4 & \multicolumn{1}{|c}{$94.83 \pm 0.64$} & \multicolumn{1}{c}{$97.49 \pm 0.80$} & \multicolumn{1}{|c}{$95.26 \pm 0.29$} & \multicolumn{1}{c}{$84.70 \pm 0.44$} & \multicolumn{1}{|c}{$96.69 \pm 0.18$} & \multicolumn{1}{c}{$95.84 \pm 0.18$} \\ \hline
\end{tabular}}
\vspace{-0.5cm}
\label{tab: height}
\end{table*}

\subsection{Sensitivity Analysis}
In this subsection, we perform a sensitivity analysis on two essential parameters in the SIDM framework: the similarity metric in state abstraction and the encoding tree height in action abstraction.

In the single-agent scenario, we investigate the bisimulation metric \citep{castro2020scalable}, successor feature similarity \citep{hoang2021successor}, Hilbert representation difference \citep{park2024foundation}, and contrastive representation similarity \citep{eysenbach2022contrastive} as alternative approaches for quantifying state similarity in state abstraction, resulting in the SISA variants SISA${\text{bm}}$, SISA${\text{sfs}}$, SISA${\text{hrd}}$, and SISA${\text{crs}}$, respectively.
For the SISA$_{\text{sfs}}$ variant, we extract temporally aligned state-action pairs to learn their successor features, using the inner product of the resulting feature representations to quantify similarity.
For the SISA$_{\text{hrd}}$ variant, we calculate the ratio between the Hilbert representation difference and the maximum observed difference across the dataset as the state similarity.
For the SISA$_{\text{crs}}$ variant, we define positive and negative pairs for contrastive learning based on state community partitions to obtain state representations, and again use the inner product of these representations as the similarity measure.
We evaluate the SISA mechanism and its variants on four DMControl tasks and summarize their average task rewards after convergence in Table \ref{tab: similarity}.
Across different similarity metrics, our state abstraction mechanism consistently achieves stable decision-making performance, demonstrating the generalization ability of our method and justifying the use of a simple similarity measure based on the Pearson correlation coefficient.

In the multi-agent collaboration setting, we adjust the encoding tree height parameter $K$ in the action abstraction to $3$ and $4$. 
For each encoding tree, abstract actions are represented as the parent nodes of leaf nodes.
We summarize SIRD's performance at different tree heights across four SMAC benchmark maps of varying difficulty in Table \ref{tab: height}.
The SIRD achieves the best performance at $K=2$ in both relatively easy and hard tasks. 
This is because, in these scenarios, an overly complex role division is unnecessary, and a two-layer action abstraction is sufficient to fulfill task requirements. 
Increasing the tree height expands the candidate role set, making role-based learning more challenging. 
For super-hard tasks, deeper encoding tree structures better facilitate complex agent collaboration in these extremely difficult scenarios.

\section{Conclusion} \label{section: conclusion}
This paper proposes SIDM, a novel hierarchical learning framework designed to reduce reliance on prior knowledge and manual definitions in both skill-based and role-based learning. 
It also addresses undirected limitations of current structural information principles.

Specifically, SIDM introduces an adaptive abstraction mechanism that extracts abstract state and action representations from historical interaction trajectories. 
Directed structural entropy is formally defined and optimized to capture transition dynamics between abstract states, enabling the discovery of hierarchical skills. 
Building on these foundations, we develop skill-based learning methods for single-agent decision-making and role-based strategies for multi-agent collaboration, resulting in substantial performance improvements.

Comprehensive evaluations on challenging benchmarks demonstrate that SIDM significantly and consistently outperforms state-of-the-art baselines in terms of learning effectiveness, stability, and efficiency. These comparative results highlight the importance of adaptively balancing the compression of irrelevant information with the retention of essential features, as well as dynamically capturing the temporal hierarchy of skills and roles to enhance hierarchical decision-making.
Furthermore, sensitivity analysis and ablation studies emphasize the generalizability of the SIDM framework and the individual contributions of the abstraction mechanism and direct entropy optimization.

Our objective is to provide an innovative and unified perspective on leveraging structural information in state-action trajectories to uncover hierarchical learning structures, ultimately improving decision-making performance. In future work, we plan to explore deeper encoding trees for hierarchical state-action abstraction and evaluate SIDM in more complex environments.

% \blindmathpaper

% Here is a citation \cite{chow:68}.

% Acknowledgements and Disclosure of Funding should go at the end, before appendices and references

\acks{The corresponding authors are Hao Peng and Angsheng Li. This work has been supported by NSFC through grants 62322202, 62441612 and 62432006, Local Science and Technology Development Fund of Hebei Province Guided by the Central Government of China through grant 246Z0102G, the "Pioneer” and “Leading Goose” R\&D Program of Zhejiang" through grant 2025C02044, National Key Laboratory under grant 241-HF-D07-01, and Hebei Natural Science Foundation through grant F2024210008.}

% Manual newpage inserted to improve layout of sample file - not
% needed in general before appendices/bibliography.

\newpage

\appendix
\section{Framework Details} \label{app: framework details}

\subsection{Notations} \label{app: notations}
\begin{table}[h]
    \centering
    \caption{Glossary of Notations.}
    \begin{tabular}{@{}l|l@{}}
    \toprule  
      %\hline\hline
      \textbf{Notation} & \textbf{Description}\\
      \hline
        $\mathcal{M}_s;\mathcal{M}_m;\mathcal{M}_\phi$ & Single-agent, Multi-agent, and Abstract Markov decision processes \\
        $n;n_i;\mathcal{N}$ & Batch Size; Single agent; Agent set \\
        $\mathcal{S};\mathcal{A};\mathcal{Z}$ & State space; Action space; Abstract space\\
        $S;A;Z$ & State variable; Action variable; Abstract Variable\\
        $s;a;z$ & Single state; Single action; Single abstract element\\
        $r;\mathcal{R};\mathcal{R}_\phi;\mathcal{R}^{in}$ & Reward; Reward function; Abstract reward function; Intrinsic reward function \\
        $\mathcal{P};\mathcal{P}_\phi$ & Transition function; Abstract transition function\\
        $\gamma$ & Discount factor\\
        $\pi;\pi_\rho;\pi_\kappa;\pi^h$ & Agent policy; Role policy; Skill policy; High-level policy\\
        $\rho;\Psi;t_i$ & Single role; Role space; Subtask \\
        $\kappa;\mathcal{K}$ & Single skill/option; Skill space \\
        $\tau$ & Individual action-observation history\\
        $\mathcal{I};\mathcal{T}$ & Initiation set; Termination condition\\ 
    \bottomrule
        $G;G_{dir}$ & Homogeneous weighted undirected and directed graphs \\
        $G_s;G_a;G^*$ & State graph; Action graph; Sparse graph\\
        $v;d_v;V$ & Single vertex; Vertex degree; Vertex set\\
        $e;E;E_{dir}$ & Single edge; Set of undirected edges; Set of directed edges\\
        $w;W;W_{dir}$ & Edge weight; Weight functions for undirected and directed edges\\
    \bottomrule
        $\lambda; \alpha; \nu; T$ & Root node; Tree node; Leaf node; Encoding tree\\
        $V_\alpha; \mathcal{V}$ & Vertex subset; Volume term\\ 
        $H$ & Structural entropy\\
        $L;K$ & Number of children node; Maximal height of encoding tree\\
        $\eta; U_i$ & Optimization operator; Node set locating specific layer\\
    \bottomrule
        $f; \mathcal{C}$ & Embedding function; State or action correlation\\
        $h_s; h_\alpha; h_T; h$ & State representation; Node representation; Tree height; Skill parameter\\
        $\pi_s;\pi_e$ & Stationary distribution; Eigenvector\\
        $\mu_{mg}; \mu_{cb}; \mu$ & Added nodes by \textit{merge} and \textit{combine} operations; Average value\\
        $\text{k};k$ & Filtration parameter; Index variable\\
        $\mathcal{B}; \mathcal{L}$ & Replay buffer; Training loss\\
    \bottomrule
    \end{tabular}
\end{table}

\newpage
\section{Detailed Derivations}

\subsection{Derivations of Directed Structural Entropy Variations} \label{app: der_se_var}
Because of the properties of the encoding tree $T_{dir}$, for each non-leaf node $\alpha$, it holds for its child nodes' corresponding vertex sets that:
\begin{equation}
    \bigcap_{i=1}^{L_\alpha} V_{\alpha_i} = \emptyset\text{,}\quad \bigcup_{i=1}^{L_\alpha} V_{\alpha_i} = V_\alpha\text{.}
\end{equation}
Equations \ref{equ: v_term} and \ref{equ: g_term} can be rewritten as follows:
\begin{equation}
    \begin{aligned}
        \mathcal{V}_\alpha &= \sum_{v_i \in V} \sum_{v_j \in V_\alpha} \left[\pi_s(v_i) \cdot W^\prime_{dir}(v_i,v_j)\right] \\
        &= \sum_{k=1}^{L_\alpha} \left[ \sum_{v_i \in V} \sum_{v_j \in V_{\alpha_k}} \left[\pi_s(v_i) \cdot W^\prime_{dir}(v_i,v_j)\right] \right] \\
        &= \sum_{k=1}^{L_\alpha} \mathcal{V}_{\alpha_k} \text{,}
    \end{aligned}
\end{equation}
\begin{equation}
    \begin{aligned}
        g_\alpha &= \sum_{v_i \notin V_\alpha} \sum_{v_j \in V_\alpha} \left[\pi_s(v_i) \cdot W^\prime_{dir}(v_i,v_j)\right] \\
        &= \sum_{k=1}^{L_\alpha} \left[ \sum_{v_i \notin V_\alpha} \sum_{v_j \in V_{\alpha_k}} \left[\pi_s(v_i) \cdot W^\prime_{dir}(v_i,v_j)\right] \right] \\
        &= \sum_{k=1}^{L_\alpha} \left[ \sum_{v_i \notin V_{\alpha_k}} \sum_{v_j \in V_{\alpha_k}} \left[\pi_s(v_i) \cdot W^\prime_{dir}(v_i,v_j)\right] - \sum_{l=1,l \neq k}^{L_\alpha} \left[\sum_{v_i \in V_{\alpha_l}} \sum_{v_j \in V_{\alpha_k}} \left[\pi_s(v_i) \cdot W^\prime_{dir}(v_i,v_j)\right] \right] \right] \\
        &= \sum_{k=1}^{L_\alpha} \left[ g_{\alpha_k} - \sum_{l=1,l \neq k}^{L_\alpha} g_{\alpha_l, \alpha_k}\right] \\
        &= \sum_{i=1}^{L_\alpha} g_{\alpha_i} - \sum_{i \neq j}^{L_\alpha} g_{\alpha_i, \alpha_j}\text{.}
    \end{aligned}
\end{equation}

\begin{figure}[t]
    \centering
    \includegraphics[width=\textwidth]{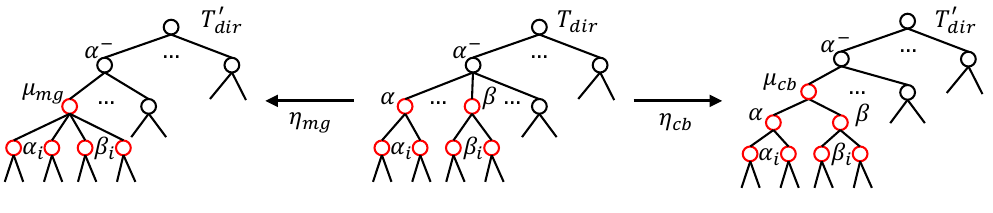}
    \caption{The \textit{merge} and \textit{combine} operations on sliding tree nodes.}\label{fig: merge combine}
\end{figure}

As shown in Figure \ref{fig: merge combine}, one \textit{merge} ($\eta_{mg}$) operation on non-leaf sibling nodes $\alpha$ and $\beta$ is executed as follows:
\begin{equation}
    \begin{cases}
        \mu_{mg}^- = \alpha^- = \beta^- \text{,}\\
        {\alpha_i}^- = \mu_{mg} & 1 \leq i \leq L_\alpha \text{,} \\
        {\beta_i}^- = \mu_{mg} & 1 \leq i \leq L_\beta \text{,}
    \end{cases}
\end{equation}
where $\mu_{mg}$ is the added tree node via the \textit{merge} operation.
Before the \textit{merge} operation, the entropy sum of nodes $\alpha$, $\beta$, and their child nodes is:
\begin{equation}
    H^{T_{dir}}(G^\prime_{dir};\alpha) + H^{T_{dir}}(G^\prime_{dir};\beta) = -\frac{g_\alpha}{\operatorname{vol}(G^\prime_{dir})} \cdot \log_2 \frac{\mathcal{V}_\alpha}{\mathcal{V}_{\alpha^-}} - \frac{g_\beta}{\operatorname{vol}(G^\prime_{dir})} \cdot \log_2 \frac{\mathcal{V}_\beta}{\mathcal{V}_{\beta^-}} \text{,}
\end{equation}
\begin{equation}
    \sum_{i}^{L_\alpha} H^{T_{dir}}(G^\prime_{dir};\alpha_i) + \sum_{i}^{L_\beta} H^{T_{dir}}(G^\prime_{dir};\beta_i) = \sum_{i}^{L_\alpha} \left[ -\frac{g_{\alpha_i}}{\operatorname{vol}(G^\prime_{dir})} \cdot \log_2 \frac{\mathcal{V}_{\alpha_i}}{\mathcal{V}_{\alpha}} \right] + \sum_{i}^{L_\beta} \left[ -\frac{g_{\beta_i}}{\operatorname{vol}(G^\prime_{dir})} \cdot \log_2 \frac{\mathcal{V}_{\beta_i}}{\mathcal{V}_{\beta}} \right] \text{.}
\end{equation}
After the \textit{merge} operation, their entropy sum is given by:
\begin{equation}
    H^{T^\prime_{dir}}(G^\prime_{dir};\mu_{mg}) = -\frac{g_{\mu_{mg}}}{\operatorname{vol}(G^\prime_{dir})} \cdot \log_2 \frac{\mathcal{V}_{\mu_{mg}}}{\mathcal{V}_{\alpha^-}} \text{,}
\end{equation}
\begin{equation}
    \sum_{i}^{L_\alpha} H^{T^\prime_{dir}}(G^\prime_{dir};\alpha_i) + \sum_{i}^{L_\beta} H^{T^\prime_{dir}}(G^\prime_{dir};\beta_i) = \sum_{i}^{L_\alpha} \left[ -\frac{g_{\alpha_i}}{\operatorname{vol}(G^\prime_{dir})} \cdot \log_2 \frac{\mathcal{V}_{\alpha_i}}{\mathcal{V}_{\mu_{mg}}} \right] + \sum_{i}^{L_\beta} \left[ -\frac{g_{\beta_i}}{\operatorname{vol}(G^\prime_{dir})} \cdot \log_2 \frac{\mathcal{V}_{\beta_i}}{\mathcal{V}_{\mu_{mg}}} \right] \text{.}
\end{equation}
Therefore, the entropy variation $\Delta_{mg}(T_{dir}, \alpha, \beta)$ is calculated as follows:
\begin{equation}
   \sum_{i}^{L_\alpha} H^{T_{dir}}(G^\prime_{dir};\alpha_i) - \sum_{i}^{L_\alpha} H^{T^\prime_{dir}}(G^\prime_{dir};\alpha_i) = \sum_{i}^{L_\alpha} \left[ -\frac{g_{\alpha_i}}{\operatorname{vol}(G^\prime_{dir})} \cdot \log_2 \frac{\mathcal{V}_{\mu_{mg}}}{\mathcal{V}_{\alpha}} \right] = -\frac{\sum_{i}^{L_\alpha} g_{\alpha_i}}{\operatorname{vol}(G^\prime_{dir})} \cdot \log_2 \frac{\mathcal{V}_{\mu_{mg}}}{\mathcal{V}_{\alpha}}\text{,}
\end{equation}
\begin{equation}
   \sum_{i}^{L_\beta} H^{T_{dir}}(G^\prime_{dir};\beta_i) - \sum_{i}^{L_\beta} H^{T^\prime_{dir}}(G^\prime_{dir};\beta_i) = \sum_{i}^{L_\beta} \left[ -\frac{g_{\beta_i}}{\operatorname{vol}(G^\prime_{dir})} \cdot \log_2 \frac{\mathcal{V}_{\mu_{mg}}}{\mathcal{V}_{\beta}} \right] =  -\frac{ \sum_{i}^{L_\beta} g_{\beta_i}}{\operatorname{vol}(G^\prime_{dir})} \cdot \log_2 \frac{\mathcal{V}_{\mu_{mg}}}{\mathcal{V}_{\beta}}\text{,}
\end{equation}
\begin{equation}
    \begin{aligned}
        H^{T^\prime_{dir}}(G^\prime_{dir};\mu_{mg}) &= -\frac{g_{\mu_{mg}}}{\operatorname{vol}(G^\prime_{dir})} \cdot \log_2 \frac{\mathcal{V}_{\mu_{mg}}}{\mathcal{V}_{\alpha^-}} \\
        &= \frac{g_{\alpha, \beta} + g_{\beta, \alpha} - g_\alpha - g_\beta}{\operatorname{vol}(G^\prime_{dir})} \cdot \log_2 \frac{\mathcal{V}_{\mu_{mg}}}{\mathcal{V}_{\alpha^-}} \text{,}
        % &= \frac{g_{\alpha, \beta} + g_{\beta, \alpha} - (\sum_{i=1}^{L_\alpha} g_{\alpha_i} - \sum_{i \neq j}^{L_\alpha} g_{\alpha_i, \alpha_j}) - (\sum_{i=1}^{L_\beta} g_{\beta_i} - \sum_{i \neq j}^{L_\beta} g_{\beta_i, \beta_j})}{\operatorname{vol}(G_{dir})} \cdot \log_2 \frac{\mathcal{V}_{\mu_{mg}}}{\mathcal{V}_{\alpha^-}} \text{,}
    \end{aligned}
\end{equation}
\begin{equation}
    \begin{aligned}
        H^{T_{dir}}(G^\prime_{dir};\alpha) &+ H^{T_{dir}}(G^\prime_{dir};\beta) - H^{T^\prime_{dir}}(G^\prime_{dir};\mu_{mg}) = \\ \frac{g_\alpha}{\operatorname{vol}(G^\prime_{dir})} \cdot \log_2 \frac{\mathcal{V}_{\mu_{mg}}}{\mathcal{V}_{\alpha}} &+ \frac{g_\beta}{\operatorname{vol}(G^\prime_{dir})} \cdot \log_2 \frac{\mathcal{V}_{\mu_{mg}}}{\mathcal{V}_{\beta}} + \frac{g_{\alpha, \beta} + g_{\beta, \alpha}}{\operatorname{vol}(G^\prime_{dir})} \cdot \log_2 \frac{\mathcal{V}_{\alpha^-}}{\mathcal{V}_{\mu_{mg}}} \text{,}
    \end{aligned}
\end{equation}
\begin{equation}
    \begin{aligned}
        \Delta_{mg}(T_{dir},\alpha,\beta) &= \frac{g_\alpha - \sum_{i}^{L_\alpha} g_{\alpha_i}}{\operatorname{vol}(G^\prime_{dir})} \cdot \log_2 \frac{\mathcal{V}_{\mu_{mg}}}{\mathcal{V}_{\alpha}} + \frac{g_\beta - \sum_{i}^{L_\beta} g_{\beta_i}}{\operatorname{vol}(G^\prime_{dir})} \cdot \log_2 \frac{\mathcal{V}_{\mu_{mg}}}{\mathcal{V}_{\beta}} + \frac{g_{\alpha, \beta} + g_{\beta, \alpha}}{\operatorname{vol}(G^\prime_{dir})} \cdot \log_2 \frac{\mathcal{V}_{\alpha^-}}{\mathcal{V}_{\mu_{mg}}} \\
        &= \frac{g_{\alpha, \beta} + g_{\beta, \alpha}}{\operatorname{vol}(G^\prime_{dir})} \cdot \log_2 \frac{\mathcal{V}_{\alpha^-}}{\mathcal{V}_{\mu_{mg}}} - \frac{\sum_{i \neq j}^{L_\alpha} g_{\alpha_i, \alpha_j}}{\operatorname{vol}(G^\prime_{dir})} \cdot \log_2 \frac{\mathcal{V}_{\mu_{mg}}}{\mathcal{V}_{\alpha}} - \frac{\sum_{i \neq j}^{L_\beta} g_{\beta_i, \beta_j}}{\operatorname{vol}(G^\prime_{dir})} \cdot \log_2 \frac{\mathcal{V}_{\mu_{mg}}}{\mathcal{V}_{\beta}}\text{.}
    \end{aligned}
\end{equation}

On the other hand, as shown in Figure \ref{fig: merge combine}, one \textit{combine} ($\eta_{cb}$) operation on sibling nodes $\alpha$ and $\beta$ is executed as follows:
\begin{equation}
    {\mu_{cb}}^- = \alpha^-\text{,}\quad \alpha^-=\mu_{cb}\text{,}\quad \beta^-=\mu_{cb}\text{,}
\end{equation}
where $\mu_{cb}$ is the added tree node via the \textit{combine} operation.
After the \textit{combine} operation, the entropy sum of these nodes is given by:
\begin{equation}
    H^{T^\prime_{dir}}(G^\prime_{dir};\mu_{cb}) = -\frac{g_{\mu_{cb}}}{\operatorname{vol}(G^\prime_{dir})} \cdot \log_2 \frac{\mathcal{V}_{\mu_{cb}}}{\mathcal{V}_{\alpha^-}} \text{,}
\end{equation}
\begin{equation}
    H^{T^\prime_{dir}}(G^\prime_{dir};\alpha) + H^{T^\prime_{dir}}(G^\prime_{dir};\beta) = -\frac{g_\alpha}{\operatorname{vol}(G^\prime_{dir})} \cdot \log_2 \frac{\mathcal{V}_\alpha}{\mathcal{V}_{\mu_{cb}}} - \frac{g_\beta}{\operatorname{vol}(G^\prime_{dir})} \cdot \log_2 \frac{\mathcal{V}_\beta}{\mathcal{V}_{\mu_{cb}}} \text{,}
\end{equation}
\begin{equation}
    \sum_{i}^{L_\alpha} H^{T^\prime_{dir}}(G^\prime_{dir};\alpha_i) + \sum_{i}^{L_\beta} H^{T^\prime_{dir}}(G^\prime_{dir};\beta_i) = \sum_{i}^{L_\alpha} H^{T_{dir}}(G^\prime_{dir};\alpha_i) + \sum_{i}^{L_\beta} H^{T_{dir}}(G^\prime_{dir};\beta_i) \text{.}
\end{equation}
Therefore, the entropy variation $\Delta_{cb}(T_{dir}, \alpha, \beta)$ is calculated as follows:
\begin{equation}
    \begin{aligned}
        \Delta_{cb}(T_{dir},\alpha,\beta) &= \left[H^{T_{dir}}(G^\prime_{dir};\alpha) + H^{T_{dir}}(G^\prime_{dir};\beta)\right] - \left[H^{T^\prime_{dir}}(G^\prime_{dir};\mu_{cb}) + H^{T^\prime_{dir}}(G^\prime_{dir};\alpha) + H^{T^\prime_{dir}}(G^\prime_{dir};\beta)\right] \\
        &= \frac{g_\alpha}{\operatorname{vol}(G^\prime_{dir})} \cdot \log_2 \frac{\mathcal{V}_{\alpha^-}}{\mathcal{V}_{\mu_{cb}}} + \frac{g_\beta}{\operatorname{vol}(G^\prime_{dir})} \cdot \log_2 \frac{\mathcal{V}_{\alpha^-}}{\mathcal{V}_{\mu_{cb}}} - \frac{g_{\mu_{cb}}}{\operatorname{vol}(G^\prime_{dir})} \cdot \log_2 \frac{\mathcal{V}_{\alpha^-}}{\mathcal{V}_{\mu_{cb}}} \\
        &= \frac{g_\alpha + g_\beta - g_{\mu_{cb}}}{\operatorname{vol}(G^\prime_{dir})} \cdot \log_2 \frac{\mathcal{V}_{\alpha^-}}{\mathcal{V}_{\mu_{cb}}} \\
        &= \frac{g_{\alpha,\beta} + g_{\beta,\alpha}}{\operatorname{vol}(G^\prime_{dir})} \cdot \log_2 \frac{\mathcal{V}_{\alpha^-}}{\mathcal{V}_{\mu_{cb}}} \text{.}
    \end{aligned}
\end{equation}

\newpage
\section{Theoretical Proofs}

\subsection{Proof of Proposition \ref{theorem: std}} \label{app: std proof}
\begin{proof}
We provide two distinct proofs for this theorem, one using the \textbf{Perron-Frobenius theorem} and another using the \textbf{Markov Chain fundamental theorem}.

\subsubsection{Proof Based on Perron-Frobenius Theorem}
For this non-negative and irreducible matrix $A_{dir}^\prime$ where each row sums to $1$, the Perron-Fronbenius theorem ensures:
\begin{itemize}
    \item There exists an eigenvalue $1$ and a corresponding eigenvector $\pi_e$ such that:
    \begin{equation}
        A_{dir}^\prime \pi_e = \pi_e \text{.}
    \end{equation}
    \item The eigenvalue $1$ is simple (i.e., it has algebraic multiplicity $1$).
    \item The corresponding eigenvector $\pi_e$ can be chosen to have strictly positive components.
    \item The absolute value of any other eigenvalue is less than $1$.
\end{itemize}
Since the eigenvalue $1$ is simple and the corresponding eigenvector $\pi_e$ has strictly positive components, we can derive the unique stationary distribution $\pi_s$ by:
\begin{equation}
    \pi_s = \frac{\pi_e}{\sum_{v \in V} \pi_e(v)} \text{.}        
\end{equation}
To verify that $\pi_s$ is indeed a stationary distribution, we need to show that it remains unchanged by the application of $A_{dir}^\prime$:
\begin{equation}
    A_{dir}^\prime \pi_s = A_{dir}^\prime \frac{\pi_e}{\sum_{v \in V} \pi_e(v)} = \frac{A_{dir}^\prime \pi_e}{\sum_{v \in V} \pi_e(v)} = \frac{\pi_e}{\sum_{v \in V} \pi_e(v)} = \pi_s \text{.}
\end{equation}
Because $G_{dir}^\prime$ is strongly connected and $A_{dir}^\prime$ is a stochastic matrix, any initial distribution $\pi^{(0)}$ will converge to the stationary distribution $\pi_s$ under repeated application of $A_{dir}^\prime$:
\begin{equation}
    \lim_{k \to \infty} (\pi^{(0)} {(A_{dir}^\prime)}^k) = \pi_s \text{.}
\end{equation}
This convergence is guaranteed by the fact that all other eigenvalues of $A_{dir}^\prime$ have magnitudes less than $1$, leading to their contributions diminishing to zero as $ k $ increases.

Therefore, the stationary distribution $\pi_s$ exists and is unique, and it corresponds to the eigenvector associated with the eigenvalue $1$ of the adjacency matrix $A_{dir}^\prime$.

\subsubsection{Proof Based on Markov Chain Theorem}
We interpret the weighted adjacency matrix $A^\prime_{dir}$ as the transition probability matrix $P$ of a Markov chain. Since each row of $A^\prime_{dir}$ sums to $1$, it satisfies the conditions of a right stochastic matrix.

Because the graph $G^\prime_{dir}$ is strongly connected, the Markov chain is irreducible, meaning that there exists a path between any two states in the chain. Additionally, since all transition probabilities are positive, the chain is aperiodic, ensuring that no strict cycle behavior prevents convergence to a unique stationary distribution.

By the fundamental theorem of Markov chains, an irreducible and aperiodic Markov chain with a finite state space has a unique stationary distribution $\pi_s$, which satisfies:

\begin{equation}
    \pi_s P = \pi_s.
\end{equation}

Expanding this equation using the transition matrix $P = A^\prime_{dir}$, we get:

\begin{equation}
    \pi_s A^\prime_{dir} = \pi_s.
\end{equation}

Thus, $\pi_s$ is a left eigenvector of $A^\prime_{dir}$ associated with the eigenvalue $1$. 

To ensure that $\pi_s$ represents a valid probability distribution, we normalize it so that:

\begin{equation}
    \sum_{v \in V} \pi_s(v) = 1.
\end{equation}

Moreover, the Perron-Frobenius theorem guarantees that for an irreducible, non-negative stochastic matrix $A^\prime_{dir}$, the eigenvalue $1$ is simple (having algebraic multiplicity $1$), and its corresponding eigenvector $\pi_s$ has strictly positive components.

Since the Markov chain is irreducible and aperiodic, it converges to the stationary distribution $\pi_s$ regardless of the initial distribution $\pi^{(0)}$:

\begin{equation}
    \lim_{k \to \infty} \pi^{(0)} (A^\prime_{dir})^k = \pi_s.
\end{equation}

This confirms that $\pi_s$ is the unique stationary distribution and corresponds to the unique eigenvector of $A^\prime_{dir}$ associated with eigenvalue $1$.

\end{proof}

\subsection{Proof of Theorem \ref{theorem: eigenoption}} \label{app: eigenoption proof}
\begin{proof}
    We analyze the structure of the discovered skill set $\mathcal{K}_h$ when $h = K$.
    In this case, the skill set is defined as:
    \begin{equation}
        \mathcal{K}_h = \{\langle \mathcal{I}_{\kappa_i}, \pi_{\kappa_i}, \mathcal{T}_{\kappa_i} \rangle \mid \alpha_i \in U^z_h\} = \{\langle \mathcal{I}_{\kappa_1}, \pi_{\kappa_1}, \mathcal{T}_{\kappa_1} \rangle \mid \alpha_1 = \lambda \}\text{,}
    \end{equation}
    where $\lambda$ is the root node in the optimal encoding tree $T^*_{dir}$, corresponding to the entire set $Z_s$ of abstract states.
    
    For the discovered skill $\kappa_1$, we explicitly define:
    \begin{itemize}
        \item The initiation set $\mathcal{I}_{\kappa_1}$ consists of all abstract states except the one with the highest stationary probability:
        \begin{equation}
            \mathcal{I}_{\kappa_1} = \{z_i^s \mid z_i^s \neq \arg \max_{z_j^s \in Z_s} \pi_s(z_j^s)\}\text{.}
        \end{equation}
        \item The skill policy maximizes an intrinsic reward $\mathcal{R}^{in}$ defined over state transitions:
        \begin{equation}
            \pi_{\kappa_1}^* = \arg \max_{\pi_{\kappa_1}} \mathbb{E}_{\mathcal{P}}\left[\sum \gamma^t \mathcal{R}^{in}(z^s_i, z^s_j)\right]\text{.}
        \end{equation}
        \item The termination condition is satisfied when the state with the highest probability is reached:
        \begin{equation}
            \mathcal{T}_{\kappa_1}(z_i^s) = 
            \begin{cases}
                1 & \text{if } \arg \max_{z_j^s \in Z_s} \pi_s(z_j^s) = z_i^s\text{,} \\
                0 & \text{otherwise}\text{.}
            \end{cases}
        \end{equation}
    \end{itemize}

    We now define the eigenoption $\kappa_e$ associated with the largest eigenvalue of the adjacency matrix $A_{dir}^\prime$.
    Eigenoptions are derived from the principal eigenvector $\pi_e$ of $A_{dir}^\prime$, which satisfies:
    \begin{equation}
        A^\prime_{dir} \pi_e = \pi_e\text{.}
    \end{equation}
    To learn the eigenoption's option policy, we define the intrinsic reward function $\mathcal{R}^{in}$ based on the eigenvector:
    \begin{equation}
        \mathcal{R}^{in}_e(z^s_i, z^s_j) = \pi_e(z^s_i) - \pi_e(z^s_j)\text{.}
    \end{equation}
    This ensures that the option policy moves toward the most ``important'' state in the eigenvector ranking.
    The eigenoption is entirely defined by:
    \begin{itemize}
        \item Initiation set:
        \begin{equation}
            \mathcal{I}_{\kappa_e} = \{z_i^s \mid z_i^s \neq \arg \max_{z_j^s \in Z_s} \pi_e(z_j^s)\}\text{.}
        \end{equation}
        \item Option policy:
        \begin{equation}
            \pi_{\kappa_e}^* = \arg \max_{\pi_{\kappa_e}} \mathbb{E}_{\mathcal{P}}\left[\sum \gamma^t \mathcal{R}^{in}_e(z^s_i, z^s_j)\right]\text{.}
        \end{equation}
        \item Termination condition:
        \begin{equation}
            \mathcal{T}_{\kappa_e}(z_i^s) = 
            \begin{cases}
                1 & \text{if } \arg \max_{z_j^s \in Z_s} \pi_e(z_j^s) = z_i^s\text{,} \\
                0 & \text{otherwise}\text{.}
            \end{cases}
        \end{equation}
    \end{itemize}
    
    According to Appendix \ref{app: std proof}, for the irreducible and stochastic matrix $A^\prime_{dir}$, we know:
    \begin{equation}
        \pi_s(z^s_i) = \pi_e(z^s_i)\text{,}
    \end{equation}
    \begin{equation}
        \mathcal{R}^{in}(z^s_i, z^s_j) = \mathcal{R}^{in}_e(z^s_i, z^s_j)\text{.}
    \end{equation}
    Thus, since both $\kappa_1$ and $\kappa_e$ share the same initiation set, option policy, and termination condition, we conclude:
    \begin{equation}
        \kappa_1 = \kappa_e\text{.}
    \end{equation}
    This proves that the only skill discovered at $h = K$ is the eigenoption associated with the largest eigenvalue.
\end{proof}

\newpage
\section{Environment Details} \label{app: environment detail}

\subsection{Visual Gridworld}
The visual gridworld environment consists of a $6 \times 6$ grid, where an agent can navigate using four discrete actions: up, down, left, and right.  
Each position on the grid corresponds to a unique state, which is represented to the agent in visual form.  
Specifically, the agent's $(x, y)$ position is encoded into a one-hot image representation, where each grid cell is depicted as a $3 \times 3$ pixel patch within an $18 \times 18$ image.  
The center pixel of the corresponding patch is activated, and the image is subsequently smoothed using a truncated Gaussian kernel to produce a more realistic visual input.  
To introduce variability and challenge the agent’s perception, per-pixel noise sampled from another truncated Gaussian distribution is added to the image.

During training, a single grid position is randomly designated as the goal state for each random seed.  
The agent receives a reward of $-1$ at every timestep until it reaches the goal state, after which a new episode begins with the agent placed in a randomly selected location that is not the goal.  
This setup encourages the agent to learn efficient navigation strategies that minimize cumulative negative reward.

\subsection{DMControl Suite}
The DeepMind Control Suite \citep{tunyasuvunakool2020dm_control} is a comprehensive benchmarking platform for reinforcement learning algorithms, featuring a diverse array of continuous control tasks that simulate various physical systems.  
Each task is designed with distinct state and action spaces, as well as task-specific reward functions.  
Together, these components define the environment’s dynamics and the agent’s learning objectives, enabling rigorous evaluation and comparison of reinforcement learning algorithms on complex control problems.

The state spaces include critical physical parameters such as positions, velocities, and angular velocities, providing agents with essential information for decision-making.  
The action spaces define how agents can influence the environment, typically through torques applied to joints or forces exerted on bodies.  
Reward functions are tailored to incentivize specific behaviors, such as balancing a pole, reaching a target, walking stably, hopping forward, catching a ball, or swinging up and stabilizing a pendulum.

\subsection{Robotic Control Environment}
This benchmark includes a suite of bipedal locomotion \citep{gehring2021hierarchical} and 7-degree-of-freedom (7-DoF) robotic manipulation \citep{silver2018residual} tasks, designed to evaluate the adaptability and robustness of reinforcement learning algorithms under dynamic and variable conditions.

In the bipedal locomotion scenarios, robots are initialized at predefined positions and configurations.  
To simulate real-world uncertainties, joint positions are perturbed with noise sampled uniformly from the interval $[-0.1, 0.1]$, while joint velocities are perturbed with Gaussian noise scaled by $0.1$.  
These perturbations are consistent with those used in standard MuJoCo benchmark tasks.  
Each environment provides three distinct observation modalities: (1) proprioceptive states, which include internal robot states such as joint angles and velocities; (2) task-specific observations, representing additional sensory inputs relevant to the task; and (3) goal state measurements, which indicate the desired outcome and are primarily used to compute relative goals in low-level policy inputs.  
Episodes typically last for $1000$ interaction steps, unless terminated earlier due to entry into invalid states, as defined by the specific robot’s configuration.

The manipulation tasks, based on a $7$-DoF robotic arm, involve diverse object interaction challenges.  
Each task introduces unique physical or dynamical variations to evaluate the generalization capabilities of learned policies:

$\bullet$ Slippery Push: The robot must push a block to a target location on a low-friction surface, increasing the difficulty of precise control.

$\bullet$ Table Cleanup: A rigid tray is introduced into the environment, requiring the robot to place a block into it while navigating around new obstacles.

$\bullet$ Pyramid Stack: The robot must stack a small red block onto a larger blue block, demanding accurate positioning and stability.

$\bullet$ Complex Hook: The robot uses a hook to manipulate objects that are otherwise unreachable, with added difficulty from random object shapes and irregularities on the table surface.

Each task is episodic, with predefined step limits (e.g., 50 or 100 steps), and sparse rewards provided only upon successful task completion.

\subsection{SMAC Benchmark}
The StarCraft Multi-Agent Challenge (SMAC) \citep{samvelyan2019starcraft} is a benchmark suite designed to evaluate cooperative multi-agent reinforcement learning (MARL) algorithms.  
Built upon the StarCraft II game engine, SMAC presents a series of micromanagement scenarios where each agent controls an individual unit and operates under partial observability and decentralized execution constraints.

In SMAC, agents receive local observations that include information about nearby allies and enemies, such as relative positions, health status, and available actions.  
The discrete action space includes movement in cardinal directions, attacking specific targets, and halting.  
The environment's dynamics are governed by the underlying StarCraft II physics and game rules, introducing complexities such as terrain advantages, unit types, and attack ranges.  
Each scenario is designed to assess various aspects of cooperative behavior, such as focus fire, unit positioning, and retreat strategies.  
The reward structure typically assigns positive rewards for eliminating enemy units and negative rewards for allied losses, thereby encouraging agents to develop effective combat tactics.

\newpage
\bibliography{0_main}

\end{document}